\documentclass[10pt,twocolumn,letterpaper]{article}

\usepackage{cvpr}              

\usepackage{graphicx}
\usepackage{amsmath}
\usepackage{amssymb}
\usepackage{booktabs}
\usepackage{pifont}
\usepackage{makecell}
\usepackage{multirow}
\usepackage[noEnd]{algpseudocodex}
\usepackage{algorithm}
\usepackage{enumitem}
\usepackage[percent]{overpic}%
\usepackage{verbatim}
\usepackage{bbm}
\usepackage{footnote}
\makesavenoteenv{table}

\newcommand{\ours}{SAI3D}

\definecolor{DarkBlue}{rgb}{0.0, 0.5, 0.8}

\newcommand{\yd}[1]{{\color{black}#1}}

\definecolor{cvprblue}{rgb}{0.21,0.49,0.74}
\usepackage[pagebackref,breaklinks,colorlinks,citecolor=cvprblue]{hyperref}

\def\paperID{2907} 
\def\confName{CVPR}
\def\confYear{2024}

\title{SAI3D: Segment Any Instance in 3D Scenes}

\author{
Yingda Yin\footnotemark[1] \textsuperscript{ 1,2} \quad 
Yuzheng Liu\footnotemark[1] \textsuperscript{ 2,3} \quad
Yang Xiao\footnotemark[1] \textsuperscript{ 4}  \\
Daniel Cohen-Or \textsuperscript{5} \quad 
Jingwei Huang \textsuperscript{6} \quad 
Baoquan Chen \textsuperscript{2,3} 
\vspace{2mm} \\
\textsuperscript{1}School of Computer Science, Peking University \quad
\textsuperscript{2}National Key Lab of General AI, China \\
\textsuperscript{3}School of Intelligence Science and Technology, Peking University   \\
\textsuperscript{4}Ecole des Ponts ParisTech \quad
\textsuperscript{5}Tel-Aviv University \quad
\textsuperscript{6}Tencent 
}

\begin{document}

\maketitle

\renewcommand{\thefootnote}{\fnsymbol{footnote}}
\footnotetext[1]{Equal contribution.}

\begin{abstract}
Advancements in 3D instance segmentation have traditionally been tethered to the availability of annotated datasets, limiting their application to a narrow spectrum of object categories. 
Recent efforts have sought to harness vision-language models like CLIP for open-set semantic reasoning, yet these methods struggle to distinguish between objects 
{of the same categories}
and rely on specific prompts that are not universally applicable.
In this paper, we introduce SAI3D, a novel zero-shot 3D instance segmentation approach that synergistically leverages geometric priors and semantic cues derived from Segment Anything Model (SAM). 
Our method partitions a 3D scene into geometric primitives, which are then progressively merged into 3D instance segmentations that are consistent with the multi-view SAM masks.
Moreover, we design a hierarchical region-growing algorithm with a dynamic thresholding mechanism, which largely improves the robustness of fine-grained 3D scene parsing.
Empirical evaluations on ScanNet\yd{, Matterport3D} and the more challenging ScanNet++ datasets demonstrate the superiority of our approach. Notably, SAI3D outperforms existing open-vocabulary baselines and even surpasses fully-supervised methods in class-agnostic segmentation on ScanNet++. Our project page is at \href{https://yd-yin.github.io/SAI3D}{https://yd-yin.github.io/SAI3D}.

\end{abstract}    
\section{Introduction}
\label{sec:intro}

\begin{figure}[htbp]
\centering
\includegraphics[width=1.\linewidth]{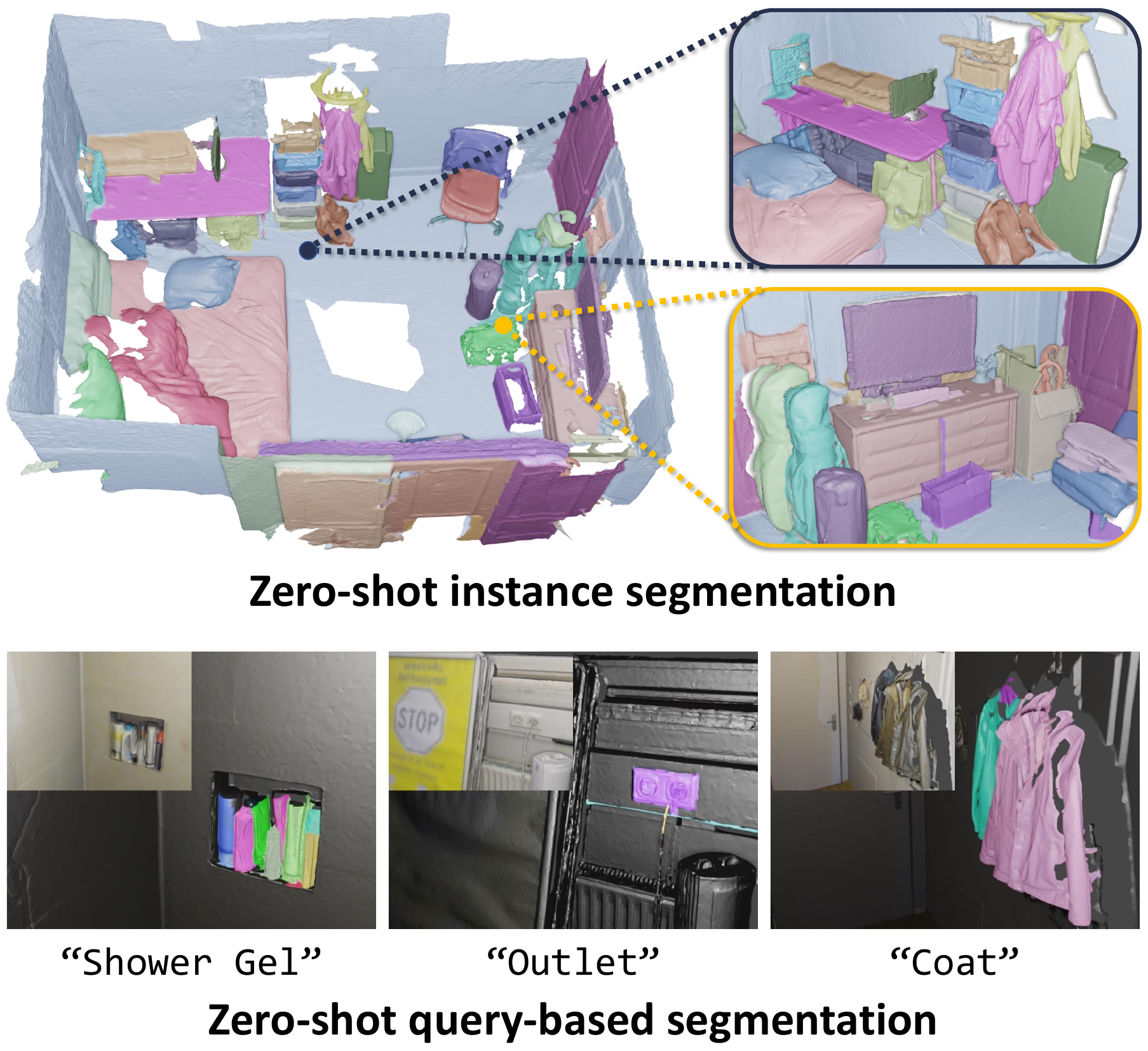}
\vspace{-4mm}
\caption{\textbf{\ours: A Zero-Shot Approach for 3D Instance Segmentation}. \ours ~ leverages  geometric priors and 2D segmentation foundation models to perform training-free zero-shot 3D instance segmentation (top). 
Our generated 3D masks enable applications of open-vocabulary queries of fine-grained 3D instances (bottom).
}
\vspace{-2mm}
\label{fig:teaser}
\end{figure}

3D instance segmentation aims to parse a 3D scene into a set of objects represented as binary foreground masks associated with semantic labels.
Although 3D instance segmentation has made great progress, state-of-the-art methods {are supervised, and} heavily rely on precise 3D annotations. Consequently, these methods are confined to a narrow scope of object categories within specific datasets like ScanNet~\cite{dai2017scannet} or KITTI~\cite{Geiger2013kitti}.
This limitation considerably constrains their applications in open-world scenarios such as embodied agents and autonomous driving.

Advanced methods, based on vision-language foundation models like CLIP~\cite{radford2021learningCLIP}, have shown impressive performance in open-set semantic reasoning.
These approaches have prompted recent studies exploring how these foundation models can assist in comprehending 3D scenes (e.g., OpenScene~\cite{Peng2023OpenScene} and LERF~\cite{kerr2023lerf}).

Although achieving open-world 3D grounding, these approaches typically predict a heatmap without distinguishing among different objects with the same semantics. Moreover, they rely on specific prompts that might not be readily available for all objects within a 3D scene.

More recently, the Segment Anything Model (SAM)~\cite{kirillov2023sam} has achieved cutting-edge results in fine-grained image segmentation within complex scenes.
Building on this progress, SA3D~\cite{cen2023SA3D} utilizes a NeRF trained on a set of 2D images. By using manual prompts in a single view (rough points representing the target object), it generates object masks within a 3D grid via cross-view self-prompting, incorporating the functionalities of SAM.
Advancing further, SAM3D~\cite{yang2023sam3d} employs SAM for automatic mask generation, partitioning images into dense instance masks, and then projecting them into the 3D scene through an iterative merging process. Notably, SAM3D is capable of generating more detailed masks compared to ground-truth annotations in ScanNet. However, it is susceptible to 2D mask errors due to the local adjacent frame merging process, which might overlook global consensus.

In this work, we investigate how to better leverage geometric priors and multi-view consistency for fine-grained instance segmentation of intricate 3D scenes.
We introduce {\ours} {(\textbf{S}egment \textbf{A}ny \textbf{I}nstance in \textbf{3D} Scenes)}, {which takes advantage of SAM to conduct zero-shot 3D instance segmentation, without training or finetuning with 3D annotations.}
{\ours} dissects the 3D scene into geometric primitives, and builds a sparse affinity matrix that captures pairwise similarity scores based on the 2D masks generated by SAM.
{Our approach involves the progressive merging of these primitives using a region-growing algorithm, and an aggregation of votes from all valid images, to obtain a multi-view consistent 3D instance segmentation.}

We evaluate our method on ScanNetV2~\cite{dai2017scannet}\yd{, ScanNet200 \cite{rozenberszki2022language}, Matterport3D \cite{Matterport3D}} as well as the recently developed ScanNet++~\cite{yeshwanth2023scannet++}. Notably, ScanNet++ provides more detailed segmentation masks with rich semantics, thereby offering a more realistic and challenging benchmark for in-the-wild scenarios.
For class-agnostic segmentation, our method significantly outperforms SAM3D and other open-vocabulary segmentation baselines on both datasets. 
Remarkably, when evaluated on ScanNet++, our zero-shot method achieves better results when compared with  \textit{fully-supervised} Mask3D models trained on ScanNet, underlining the efficacy of {\ours} in fine-grained instance segmentation of complex 3D scenes.

\noindent
Overall, our contributions are summarized as follows:
\begin{itemize}
    \item We introduce {\ours}, an efficient zero-shot 3D instance segmentation method combining geometric priors and semantic-aware image segmentation. 
    
    \item {We present a carefully designed aggregation method of 2D image masks into coherent 3D segmentations that are consistent across different views.}

    \item We demonstrate that the generated 3D masks are more accurate than previous approaches, opening new opportunities for unsupervised 3D learning.
\end{itemize}
\section{Related Work}
\label{sec:related}

\noindent  \textbf{Closed-vocabulary 3D segmentation.}
3D semantic segmentation is a long-studied topic, which aims to categorize each point in a given 3D scene with a specific semantic class \cite{qi2017pointnet,qi2017pointnet++,li2018pointcnn,wu2019pointconv,kundu2020virtual,graham20183d,hu2020randla,wang2019dynamic,rozenberszki2022language,Peng2023OpenScene,zhang2023growsp}. 
3D instance segmentation extends this by identifying distinct objects within the same semantic category and assigning unique masks to each object instance \cite{choy20194d,nekrasov2021mix3d,thomas2019kpconv,rethage2018fully,wang2018sgpn,hu2020randla,schult2023mask3d,takmaz2023openmask3d,fan2021scf,han2020occuseg,hou20193d,liang2021instance,vu2022softgroup++,hui2022learning,kolodiazhnyi2023top,sun2023superpoint}. 
ScanNet200~\cite{dai2017scannet} is a standard benchmark used for indoor 3D instance segmentation evaluation, and Mask3D~\cite{schult2023mask3d} achieves state-of-the-art performance on it using a transformer-based network.
Despite its advancements, Mask3D still requires a large amount of 3D annotated data for network training, as previous supervised learning methods.
This hampers generalizing the method towards open-world scenarios containing novel objects of unseen categories.
Moreover, the annotated 3D data is expensive to collect, and sometimes even impossible due to privacy reasons.
In this paper, we focus on zero-shot open-vocabulary 3D segmentation, where no training or finetuning with 3D annotation is needed.

\vspace{1mm}
\noindent \textbf{Open-vocabulary 2D image segmentation.}
By training on the large-scale web data of image-text pairs, foundation image-language models such as CLIP~\cite{radford2021learning} have achieved impressive performance in aligning image and text in high-dimensional feature space.
The following works apply CLIP feature to various zero-shot image tasks, including image captioning~\cite{xu2022clip,cho2022fine}, object recognition~\cite{hegde2023clip}, and object detection~\cite{lin2023gridclip,gu2021open}.
More recently, OpenSeg~\cite{ghiasi2022scaling} and OV-Seg~\cite{liang2023open} extend foundation image-language models to semantic image segmentation by learning a semantic-aware pixel-wise embedding.
Segment Anything Model (SAM)~\cite{kirillov2023sam} takes a step further to segment any object, in any image, with user-provided or automatically-generated prompts. 
SAM has learned a general notion of what objects are, which enables zero-shot generalization to unfamiliar objects and images without requiring additional training.
In our work, we benefit from the zero-shot generalization of SAM to produce high-quality 2D masks on multi-view images, and aggregate them into consistent 3D segments using a primitive-based region growing.

\vspace{1mm}
\noindent \textbf{Open-vocabulary 3D segmentation.}
Inspired by the successful 2D open-vocabulary segmentation models, OpenScene~\cite{Peng2023OpenScene} proposes an open-vocabulary 3D semantic segmentation by distilling the CLIP feature onto 3D point clouds.
On the other hand, LERF~\cite{kerr2023lerf} and DFF \cite{kobayashi2022decomposingDFF} integrate language within NeRF by optimizing an extra feature field that aligns with the CLIP feature.
While these methods enable open-vocabulary querying by text prompts, they cannot distinguish between object instances of the same categories.
To solve this issue, OpenMask3D~\cite{takmaz2023openmask3d} and OpenIns3D~\cite{huang2023openins3d} leverage the pre-trained Mask3D models for class-agnostic 3D mask proposal generation.
While achieving promising instance segmentation results on indoor scenes with similar objects as the training data (ScanNet), we show in our experiments that they fail in complex scenes with fine-grained objects.
\yd{Concurrent work, MaskClustering~\cite{yan2024maskclustering} and Open3DIS~\cite{nguyen2023open3dis}, both utilize 2D foundation models to obtain segmentation masks, subsequently aggregating these masks to construct 3D representations.}
\yd{In this work, we integrate both 3D geometric priors and semantic clues from multi-view 2D masks from SAM model for fine-grained 3D instance segmentation.}

\begin{figure*}[t]
\centering
\includegraphics[width=0.95\linewidth]{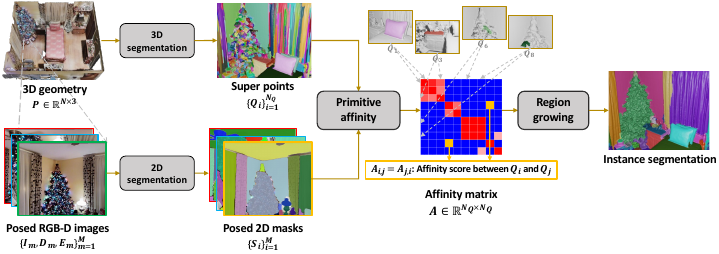}
\vspace{-2mm}
\caption{\textbf{Method overview.} 
Our approach combines geometric priors with the capabilities of 2D foundation models.
We over-segment 3D point clouds into superpoints (top-left), and generate 2D image masks using SAM (bottom-left).
We then construct a scene graph that quantifies the pairwise affinity scores of super points (middle).
Finally, we leverage a progressive region growing to gradually merge 3D superpoints into the final 3D instance segmentation masks (right).
}
\vspace{-2mm}
\label{fig:overview}
\end{figure*}

\section{Method}
\label{sec:method}

We build a training-free zero-shot 3D instance segmentation framework based on powerful 2D foundation models (e.g., SAM). 
Formally, we assume a 3D scene represented as point cloud $P \in \mathbb{R}^{N\times3}$, together with a set of posed RGB-D images $\{I_m, D_m, E_m\}_{m=1}^M$, where $I_m$ and $D_m$ denotes the RGB image and depth map, and $E_m$ is the corresponding camera extrinsic parameters.
We aim to predict a set of 3D instance masks representing different object instances in that scene.

\vspace{1mm}
\noindent \textbf{Overview.}
The overview of our approach is shown in Fig.~\ref{fig:overview}.
First, we group scene points into 3D primitives $\{Q_i\}_{i=1}^{N_Q}$ using normal-based graph cut, and predict 2D image masks $\{S_i\}_{i=1}^M$ based on the SAM automatic mask generation.
The 3D point grouping incorporates geometry information, while the 2D image segmentation inherits the powerful parsing ability from image foundation models.
Then, we build a scene graph with nodes corresponding to the 3D primitives, and each edge represents the pairwise affinity score computed based on the related 2D masks.
Finally, we obtain the 3D instance masks by merging neighboring primitives with large affinity scores, which is implemented using a progressive region growing algorithm.

\subsection{Scene Graph Construction}
\label{sec:scene_graph}

Based on the 3D scene point cloud, we group 3D points with similar geometric properties into continuous regions, and represent these regions as the scene nodes. For each node, we aggregate a set of related images as well as the corresponding 2D masks. 
We build graph edges to connect neighboring nodes, and weight each edge with the primitive similarity, which is computed by comparing two sets of image masks corresponding to the primitives.

\vspace{1mm}

\noindent \textbf{3D primitives.}
We follow recent works~\cite{rozenberszki2023unscene3d,yang2023sam3d} to group points with similar geometric properties into 3D primitives.
Specifically, we apply a normal-based graph cut algorithm \cite{felzenszwalb2004efficient} to over-segment the point cloud $P\in \mathbb{R}^{N\times 3}$ into a set of superpoints $\{Q_i\}_{i=1}^{N_Q}$. 
Compared with a scene graph built at the point-level, the transformation of unstructured 3D points to geometry-based primitives enables efficient handling of unstructured 3D data.
{More importantly, the affinity scores computed between primitives are more reliable than those computed between points, which largely improves the robustness of our approach.}

\vspace{1mm}
\noindent \textbf{2D masks.}
We employ the auto mask generation technique of SAM to obtain 2D object masks on the RGB images. 
We note that when a pixel is covered by multiple masks, only the one with the highest predicted IoU is maintained to achieve distinct, non-overlapping masks.
As shown in Fig.~\ref{fig:overview}, the predicted 2D masks represent the notion of objects learned by SAM, and we incorporate it into the 3D scenes via a primitive-based region growing.

\begin{figure}[t]
\centering
\includegraphics[width=0.9\linewidth]{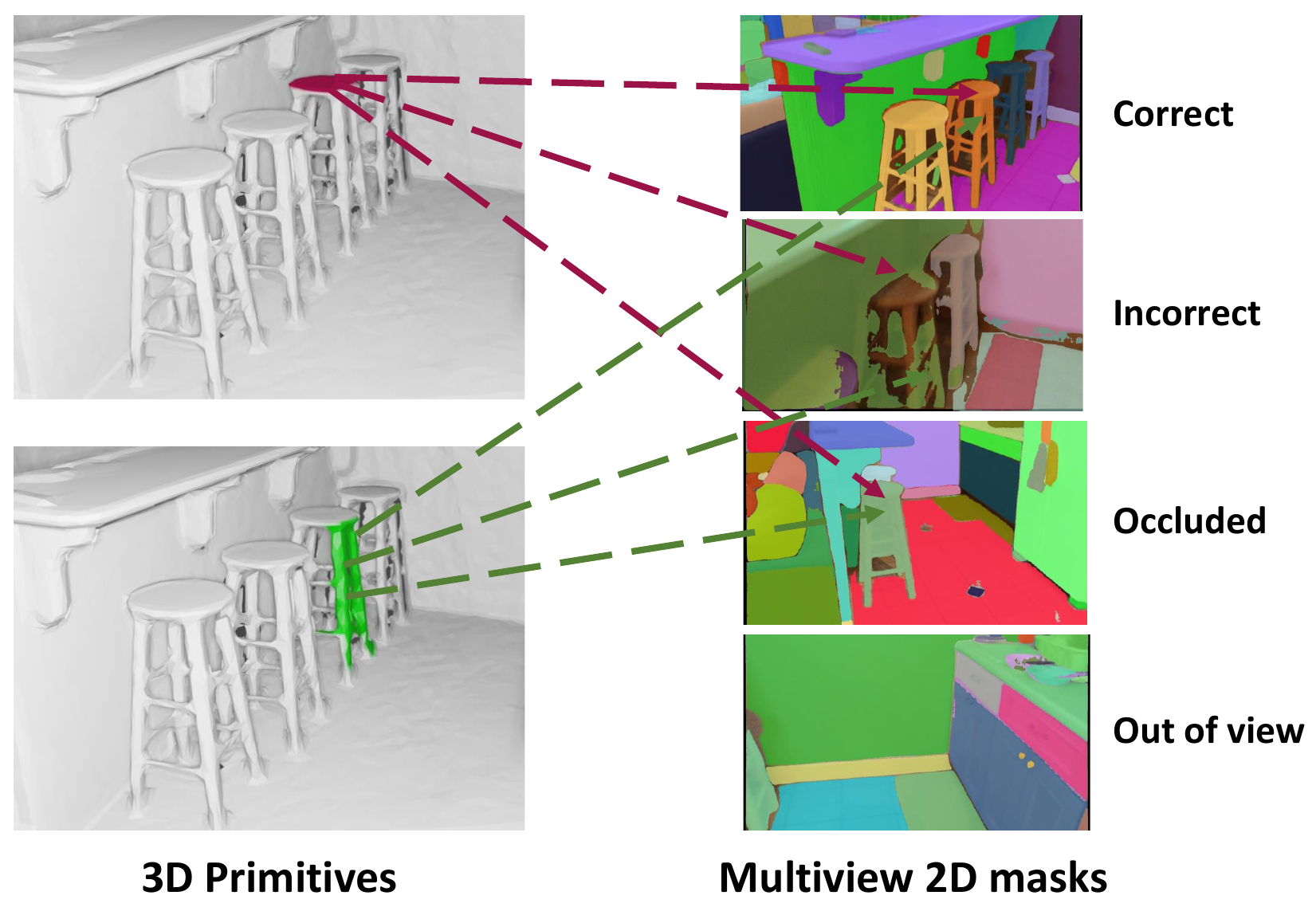}
\vspace{-1mm}
\caption{\textbf{3D-2D projections.} 
Affinity scores for 3D primitives are derived by projecting them onto multi-view 2D masks. In the provided example, accurate masks (first row) confirm object unity, like parts of a stool, while incorrect masks (second row) introduce noise in affinity assessment. Points occluded in images (third row) or outside image boundaries (fourth row) are excluded from affinity score calculations to ensure segmentation accuracy.
}
\vspace{-2mm}
\label{fig:primitive-projection}
\end{figure}

\subsection{Primitive Affinity}
\label{sec:affinity_super_points}

Given a pair of 3D primitives $Q_i, Q_j$, the affinity score $A_{i, j}$ between them represents the likelihood of their belonging to the same object instance.
We compute this affinity score by comparing the corresponding 2D masks covered by the projected 3D primitives.
This results in an adjacency matrix $A \in \mathbb{R}^{N_Q \times N_Q}$, which will be used for merging primitives with high-affinity scores.

\vspace{1mm}
\noindent \textbf{Primitive projection.}
We consider the common pinhole camera matrix for 2D-3D projection in our approach.
For the $i$-th 3D primitive $Q_i$, we obtain its projection on the $m$-th image by rendering $Q_i$ with the corresponding camera pose parameters $E_m$:
\begin{equation}\label{eq:projection}
    Q_{i, m}^\text{2D} = \Pi( Q_i, E_m)
\end{equation}
where $\Pi(\cdot)$ is the point rendering operator.
As shown in Fig.~\ref{fig:primitive-projection}, the projected primitives can be partially visible or completely occluded in the image.
We compute the visibility of a projected primitive as the ratio of 3D points that are visible in the image.
For each primitive, we discard images where the primitive visibility is zero, and keep the rest images as valid ones.

\vspace{1mm}
\noindent \textbf{Affinity in a single view.}
Based on the projected primitive $Q_{i, m}^\text{2D}$ and the image segmentation $S_m$, we collect the mask labels covered by $Q_{i, m}^\text{2D}$ and compute a normalized histogram of the mask labels \yd{as a vector, denoted as $\mathbf{h}_{i, m}$}.
This histogram represents a distribution of 2D instance mask labels corresponding to the projected primitive.
As shown in Fig.~\ref{fig:primitive-projection}, a projected primitive can have multiple 2D instance mask labels, represented in different colors.

We compute the affinity score between two primitives projected in the $m$-th image as the cosine similarity between two vectors:
\begin{equation}\label{eq:sim-single-view}
    A_{i, j}^{m} =  \frac{ \mathbf{h}_{i, m} \cdot \mathbf{h}_{j, m} }{ | \mathbf{h}_{i, m} | | \mathbf{h}_{j, m} | }.
\end{equation}

\vspace{1mm}
\noindent \textbf{Affinity in multiple views.}
Since 3D primitives are observed in different images, the affinity scores $A_{i, j}^{m}$ can also vary across different views.
We treat the affinity score of each valid as a candidate, and combine them together using a voting scheme in order to achieve cross-view consistency.
Formally, we compute a weight for each candidate $A_{i, j}^{m}$, and get the final affinity score using weighted-sum:
\begin{equation}\label{eq:affinity}
    A_{i, j} = \frac{1}{\sum_{m=1}^{M} w_{i, j}^{m}} \sum_{m=1}^{M} \left( w_{i, j}^{m} A_{i, j}^{m}
    \right).
\end{equation}
The weight $w_{i, j}^{m}$ is calculated as the product of visibilities of $Q_{i, m}^\text{2D}$ and $Q_{j, m}^\text{2D}$ \yd{as Eq. \ref{eq:weight}
\begin{equation}
    \footnotesize
    w_{i,j}^m = \frac{\sum_{p\in Q_i}\mathbbm{1}\left( \text{Valid}(p, S_m) \right)}{|Q_i|} \frac{\sum_{p\in Q_j}\mathbbm{1}\left( \text{Valid}(p, S_m) \right)}{|Q_j|} 
    \label{eq:weight}
\end{equation}
where ``Valid'' function indicates if the projected point $p$ is visible in the 2D segmentation mask $S_m$.}
Note that $w_{i, j}^{m} = 0$ for invalid images where either one of the primitives is not visible.

\begin{figure}[t]
\centering
\includegraphics[width=0.9\linewidth]{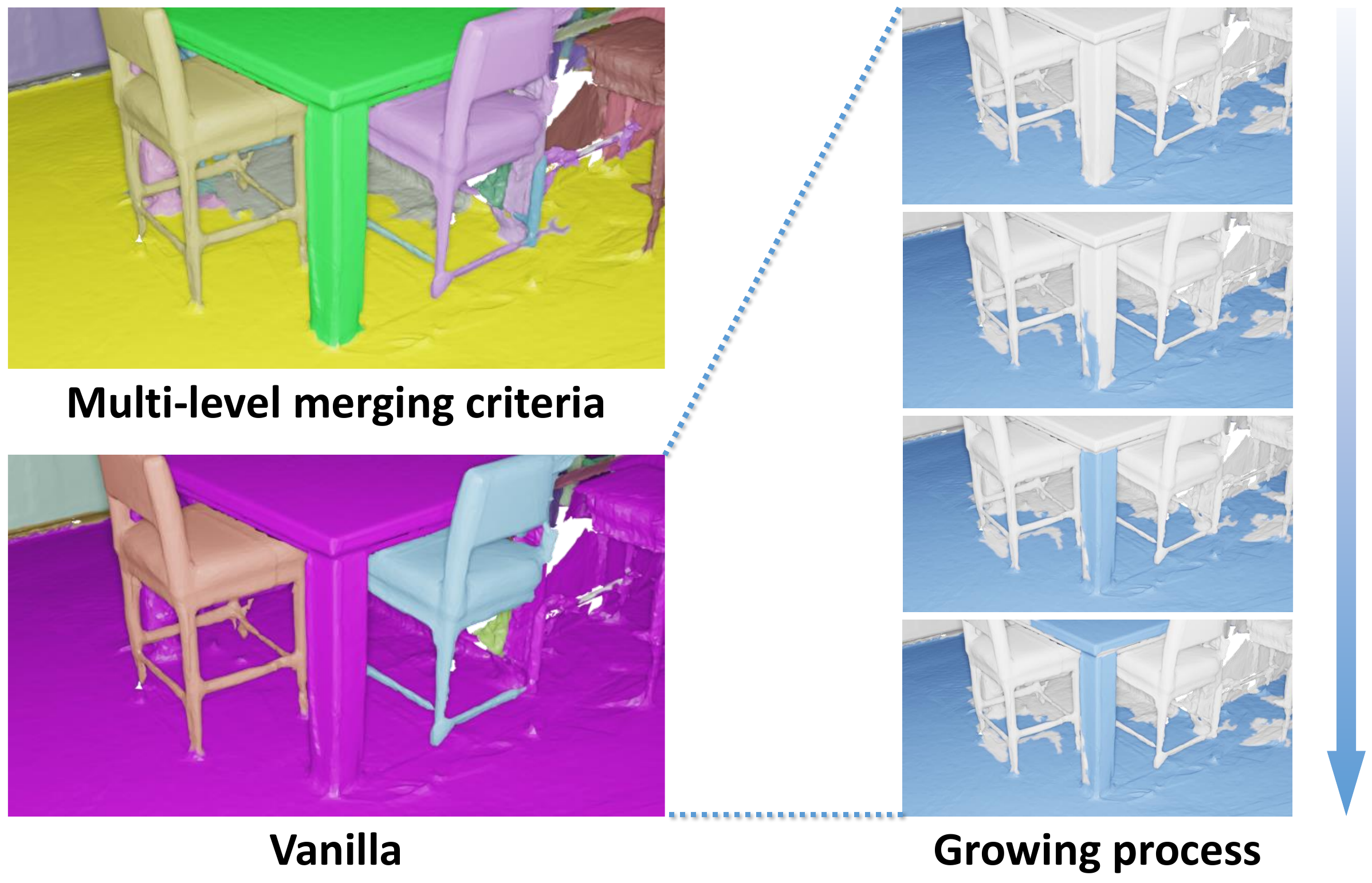}
\vspace{-2mm}
\caption{\textbf{Multi-level merging criteria}. \textbf{(Left)} Compared with the vanilla region growing that accumulates merging errors during the growing process, our approach achieves better results using multi-level merging criteria.
{\textbf{(Right)} The vanilla algorithm mistakenly merges the entire table with the ground, triggered merely by an incorrect affinity between tiny segments of the table leg and the ground.}
}
\vspace{-4mm}
\label{fig:multi-level-merging}
\end{figure}

\subsection{Primitive Merging}
\label{sec:region-growing}

Based on the scene graph and the computed affinity matrix, we obtain the final 3D instance masks using a region-growing algorithm that gradually merges 3D primitives with large affinity scores.
We design a progressive region-growing algorithm, with multi-level merging criteria.

\vspace{1mm}

\noindent \textbf{Multi-level merging criteria.}
Given a region represented as a queue of primitives, the top primitive is popped out and used to retrieve the neighboring nodes for growing.
In the vanilla region growing algorithm, a node would be added to the region if it shares a high-affinity score with the popped node. 
However, this pairwise comparison is prone to errors, which would accumulate along with the growing process, {as shown in Fig. \ref{fig:multi-level-merging}}.
To solve this problem, we propose a multi-level merging criteria that compute the affinity score in a hierarchical way, where the affinity scores between the candidate node and all the nodes within the region are summed together by weighting them according to the graph distance.

\begin{figure*}[t]
    \centering
    \begin{tabular}{cccc}
    \includegraphics[width=0.2\linewidth]{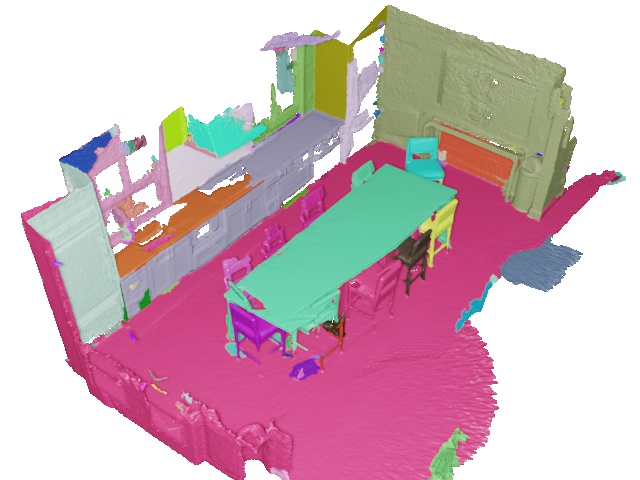} \hspace{1mm}
    &\includegraphics[width=0.2\linewidth]{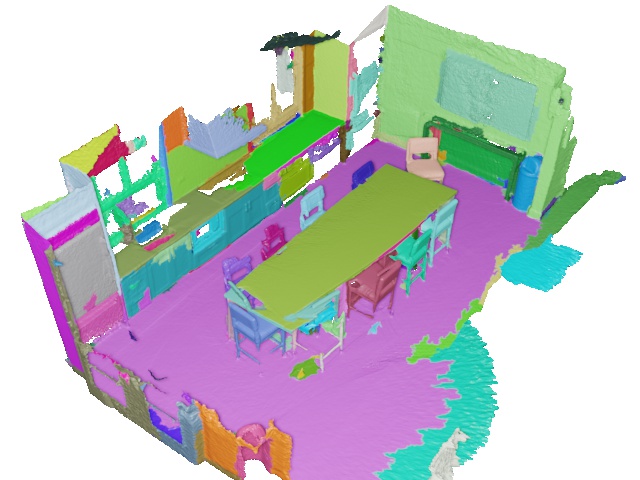} \hspace{1mm}
    &\includegraphics[width=0.2\linewidth]{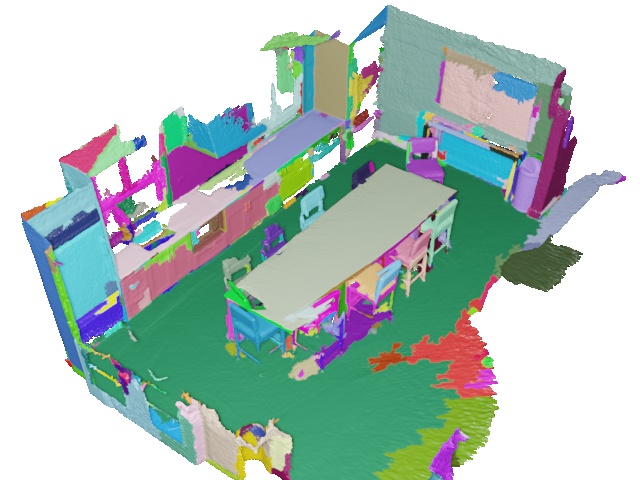}  \hspace{1mm}
    &\includegraphics[width=0.2\linewidth]{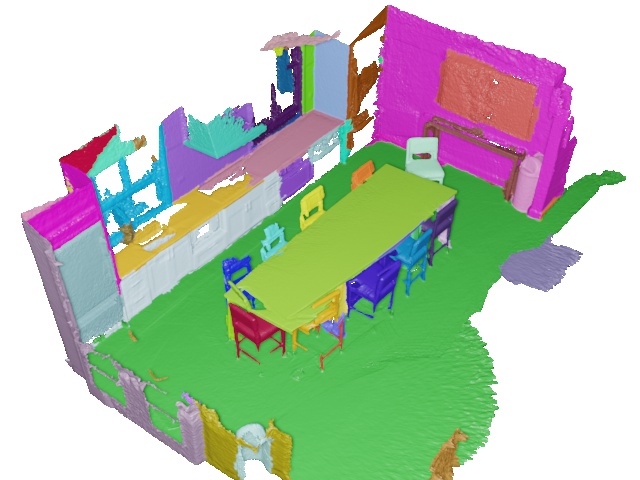} \\
    
    \includegraphics[width=0.18\linewidth]{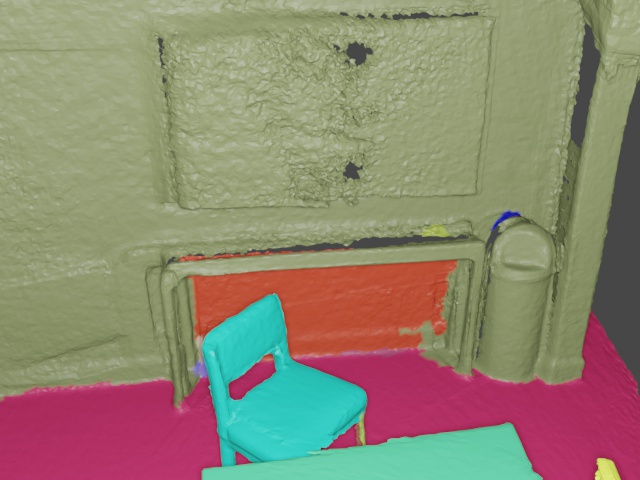}    \hspace{1mm}
    &\includegraphics[width=0.18\linewidth]{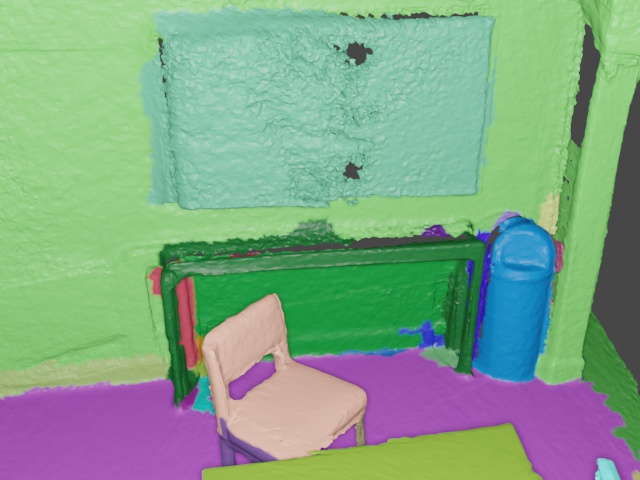}   \hspace{1mm}
    &\includegraphics[width=0.18\linewidth]{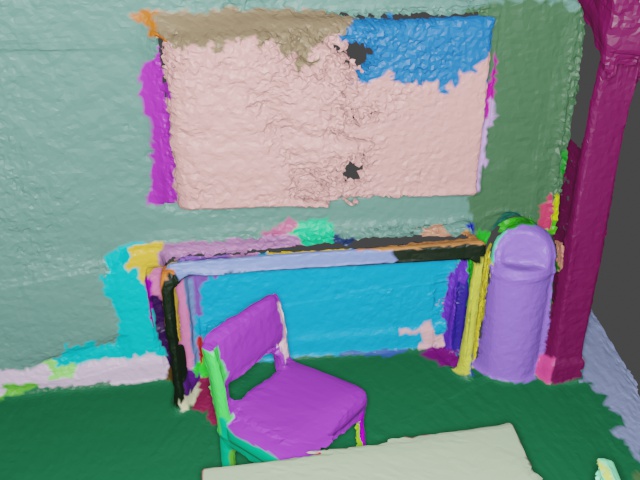}  \hspace{1mm}
    &\includegraphics[width=0.18\linewidth]{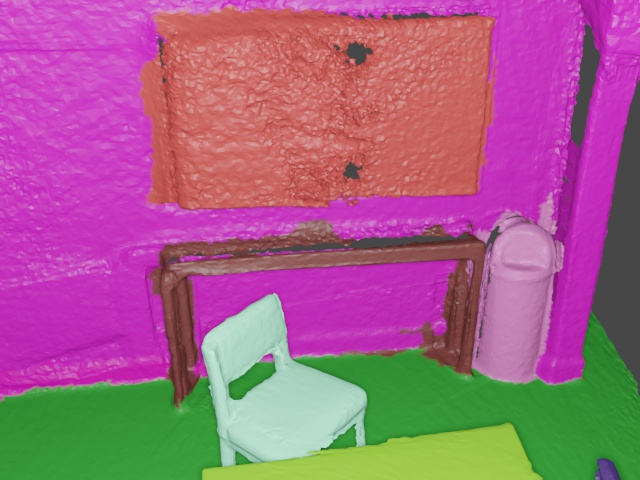} \\

    \small{Low threshold} \hspace{1mm} & \small{Middle threshold} \hspace{1mm} &\small{High threshold} \hspace{1mm} &\small{Progressive threshold}  

    \end{tabular}
    \vspace{-1mm}
    \caption{\small {\textbf{Different thresholding strategies.} 
    Without dynamic thresholding, the segmentation results can be sensitive to the manual affinity threshold. 
    A lower threshold is prone to under-segmentation, such as the merged television and the wall (first column). Conversely, a higher threshold may result in over-segmentation, breaking objects into messy parts (third column). 
    Our progressive thresholding introduces a dynamic threshold along with the merging process, thus contributing to more robust and accurate segmentation.
    }}
	\label{fig:connect_thres}
\end{figure*}

\vspace{1mm}

\noindent \textbf{Progressive growing.}
Another important hyper-parameter of region-growing is the affinity threshold, which is used to determine if two regions should be merged or not.
As shown in Fig.~\ref{fig:connect_thres}, the region-growing process is sensitive to the threshold value, and setting a fixed threshold usually results in over-segmentation or under-segmentation.
Based on this observation, we propose a progressive growing algorithm, where the growing process is decomposed into multiple stages with threshold values varying from high to low.
In this way, we merge small regions with a strict criterion at the beginning, which prevents incorrect merges from accumulating along with the growing process.
As the regions gradually grow into large ones that are more reliable, we use a more relaxed criterion to merge them.
This dynamic thresholding approach ensures that our merging framework remains sensitive to the evolving certainty of connectivity, thereby enhancing the accuracy of the final segmentation.

\subsection{Open-vocabulary 3D Object Search}

Our generated 3D instance masks enable the application of open-vocabulary queries of fine-grained 3D objects.
Given a text prompt describing the target object, we leverage the 2D segmentation method OVSeg~\cite{liang2023open} to obtain semantic masks on the 2D images.
We then back-project the 2D object masks onto 3D points, and compute the overlap between the projected masks and the generated 3D instance masks.
The 3D instances having an overlap larger than 50\% will be assigned as the target object masks.

\section{Experiment}
\label{sec:exp}

We evaluate {\ours} on multiple datasets to demonstrate its effectiveness in 3D instance segmentation.
We compare it with leading methods in open-vocabulary segmentation, which, like \ours, are designed for zero-shot transfer scenarios. Additionally, we benchmark against the state-of-the-art closed-vocabulary methods that require training on annotated datasets.

\subsection{Experiment Setup}

\noindent \textbf{Datasets.}
\textbf{ScanNet++}~\cite{yeshwanth2023scannet++} is a very recently released indoor dataset that offers posed RGB-D streams, high-quality 3D geometry captured by advanced laser technology, and comprehensive object annotations. 
Compared with previous datasets, ScanNet++ features higher-resolution 3D geometry and finer-grained data annotations, especially for long-tail semantics, which poses new challenges that reflect real-world applications.
\yd{To better validate the robustness of our method, we also incorporate the well-studied \textbf{ScanNetV2} \cite{dai2017scannet}, \textbf{ScanNet200} \cite{rozenberszki2022language} (with 200 semantic classes) and \textbf{Matterport3D}\cite{Matterport3D} datasets. Please see supplementary for more details.} 

\vspace{1mm}
\noindent \textbf{Evaluation metrics.}
We evaluate the numerical results with the widely-used Average Precision score. Following the baselines \cite{takmaz2023openmask3d,schult2023mask3d,rozenberszki2023unscene3d}, we report scores at IoU scores of 25 \% and 50 \% (AP@25, AP@50) and averaged over all overlaps between [50 \% and 95 \%] at 5 \% steps.
We adopt two evaluation setups: \textit{class-agnostic} instance segmentation focuses only on the accuracy of the instance masks themselves, 
and \textit{semantic} instance segmentation that also considers their associated semantic labels. 
We calculated the average score across all semantic categories to obtain the overall performance.

\vspace{1mm}
\noindent \textbf{Baselines.} 
We compare our approach with both closed-vocabulary and open-vocabulary baselines. 
\yd{
{Mask3D} is the state-of-the-art, transformer-based method, supervised with training data. 
For open-vocabulary methods, we compare with the recent SAM3D \cite{yang2023sam3d}, UnScene3D \cite{rozenberszki2023unscene3d}, and OVIR-3D \cite{lu2023ovir}.}
Notably, SAM3D is very similar to our approach by building upon the automatic mask generation of SAM, but differs in the merging process. Unlike our approach based on 3D primitives, SAM3D projects 2D masks onto 3D and iteratively merges them frame by frame, which overlooks the global geometry properties of the 3D scene.
{OpenMask3D} \cite{takmaz2023openmask3d} is built upon {Mask3D} with supervised mask proposals and open-vocabulary semantic assignment.
In addition, We also compare with the traditional point grouping methods like {HDBSCAN}~\cite{mcinnes2017accelerated} and {Felzenszwalb}’s algorithm~\cite{felzenszwalb2004efficient}, as well as a {feature clustering} method~\cite{nunes2022unsupervised}.

\subsection{Results}

\setlength{\tabcolsep}{7pt}
\begin{table}
    \centering
    \footnotesize
    \caption{\textbf{Class-agnostic 3D instance segmentation on ScanNet++ dataset.} 
    We compare against both closed-vocabulary and open-vocabulary methods, and report average precision scores.
    Note that Mask3D is trained on ScanNetV2 dataset.} 
    \vspace{-2mm}
    \begin{tabular}{lcccc}
    \toprule
        Method & Training Set & AP & AP$_{50}$ & AP$_{25}$ \\
        \midrule 
        \textit{Closed-vocabulary} \\
        Mask3D \cite{Schult23ICRA} & ScanNetV2 & 9.9 & 17.3 & 25.8 \\
        \midrule
        \textit{Open-vocabulary} \\
        Felzenszwalb et al. \cite{felzenszwalb2004efficient} & - & 4.1 & 9.2 & 25.3 \\
        SAM3D \cite{yang2023sam3d} & - & 7.2 & 14.2 & 29.4 \\
        Ours & - & \textbf{17.1}  & \textbf{31.1} & \textbf{49.5} \\
    \bottomrule
    \end{tabular}
    \vspace{-2mm}
    \label{tab:class_agnostic_scannetpp}
\end{table}

\noindent \textbf{Fine-grained 3D segmentation.}
\yd{Table~\ref{tab:class_agnostic_scannetpp} reports the numerical results for class-agnostic instance segmentation on ScanNet++ dataset.}
Our method outperforms prior works in all evaluation metrics. 
Not only did it surpass the unsupervised methods significantly, but it also outperformed the closed-vocabulary baseline trained on ScanNetV2. This highlights our method's ability to handle detailed objects and diverse semantic categories effectively.

\begin{figure*}[t]
    \centering
    \begin{tabular}{ccccc}
    \includegraphics[width=0.18\linewidth]  {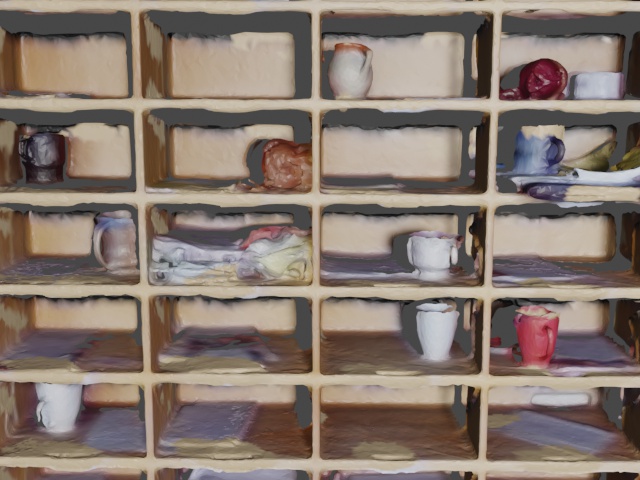} \hspace{-4mm}
    &\includegraphics[width=0.18\linewidth]{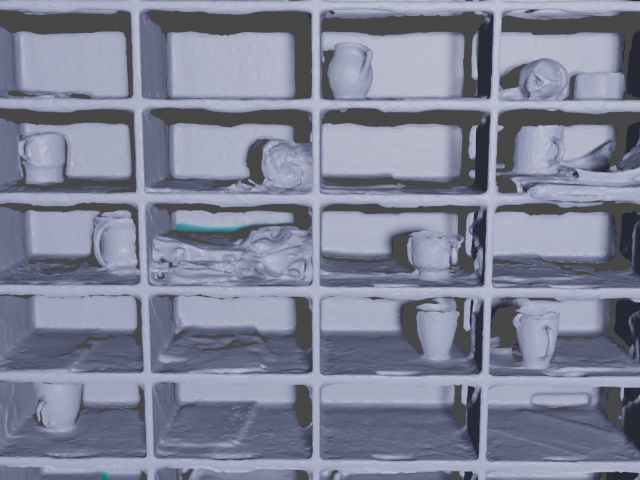}
    \hspace{-4mm}
    &\includegraphics[width=0.18\linewidth]{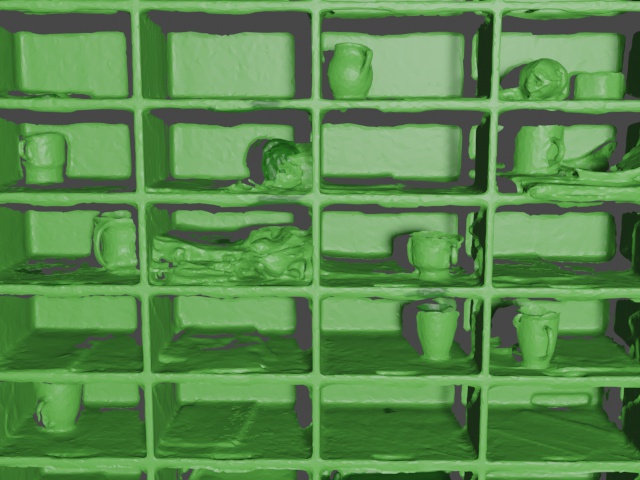}
    \hspace{-4mm}
    &\includegraphics[width=0.18\linewidth]{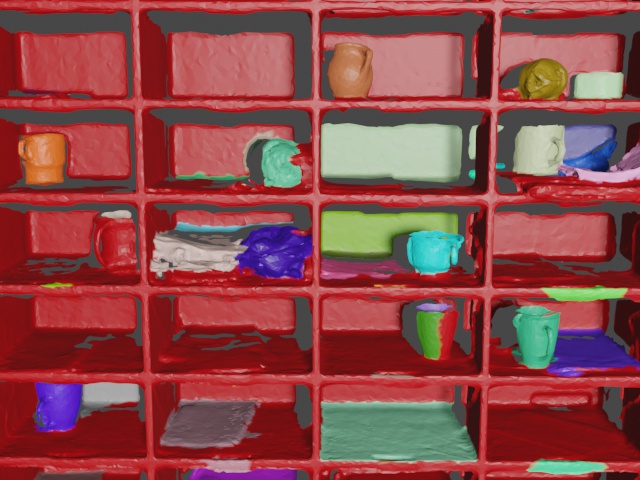}
    \hspace{-4mm}
    &\includegraphics[width=0.18\linewidth]{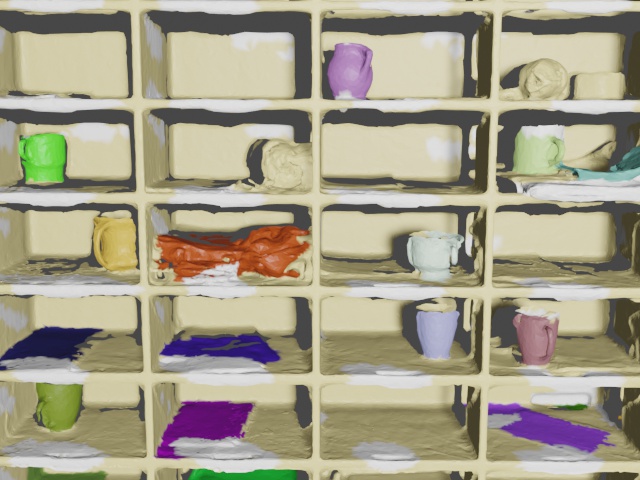} \\ 
    
    \includegraphics[width=0.18\linewidth]{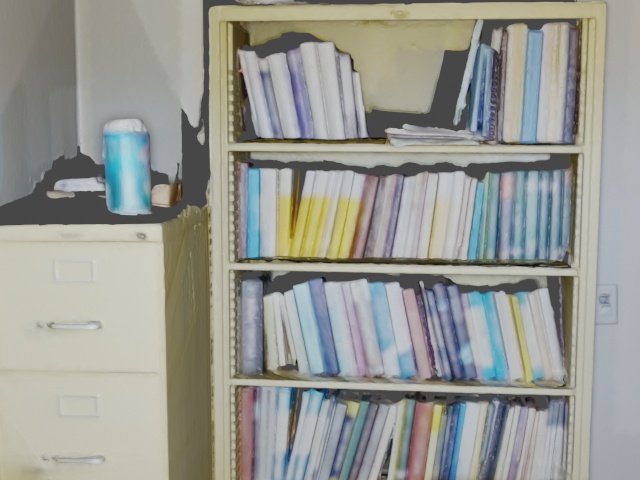}
    \hspace{-4mm}
    &\includegraphics[width=0.18\linewidth]{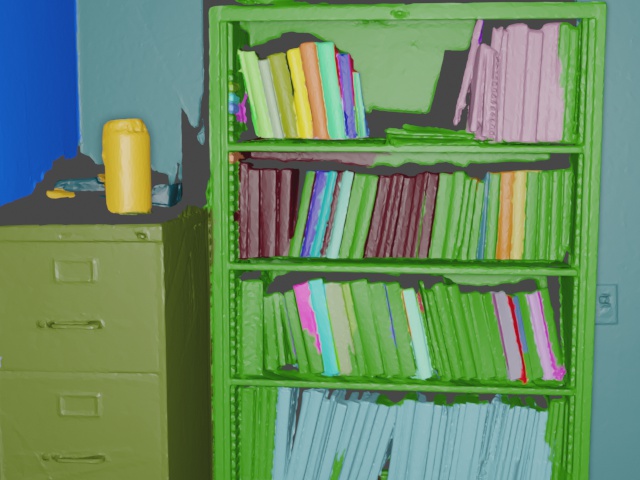}
    \hspace{-4mm}
    &\includegraphics[width=0.18\linewidth]{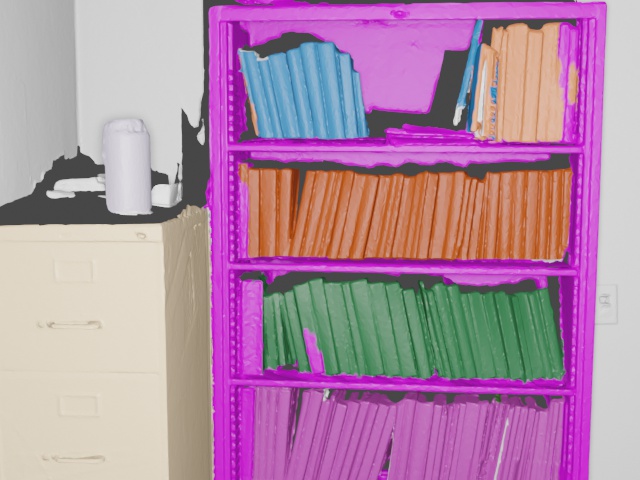}
    \hspace{-4mm}
    &\includegraphics[width=0.18\linewidth]{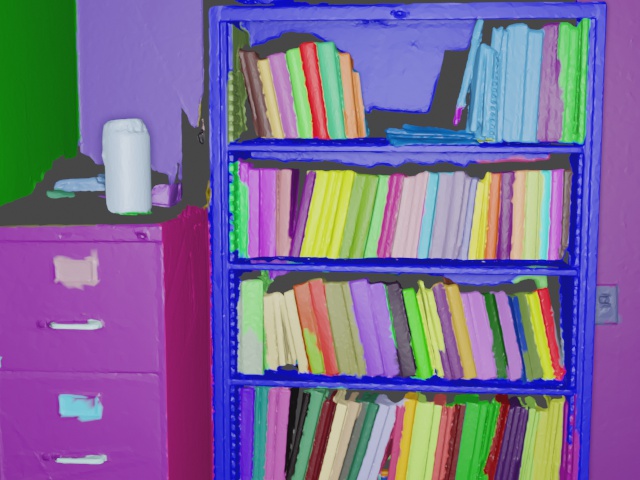}
    \hspace{-4mm}
    &\includegraphics[width=0.18\linewidth]{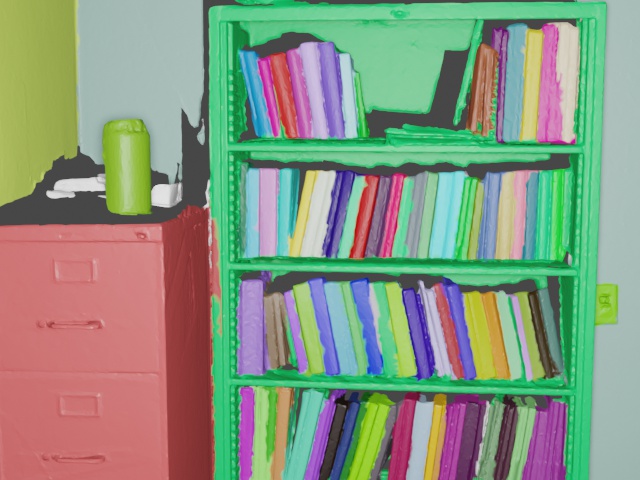} \\

    \includegraphics[width=0.18\linewidth]{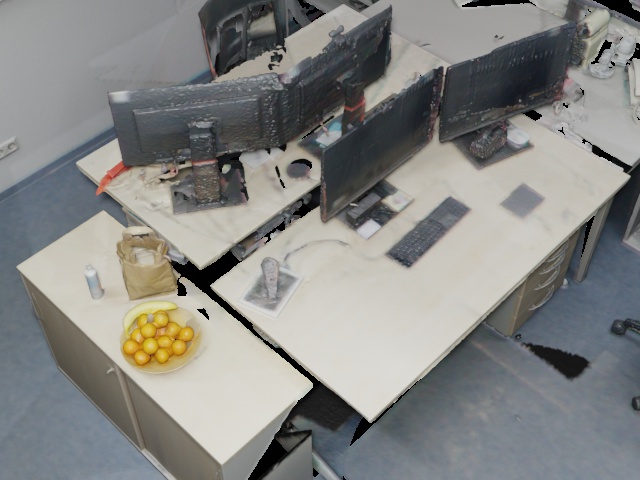}
    \hspace{-4mm}
    &\includegraphics[width=0.18\linewidth]{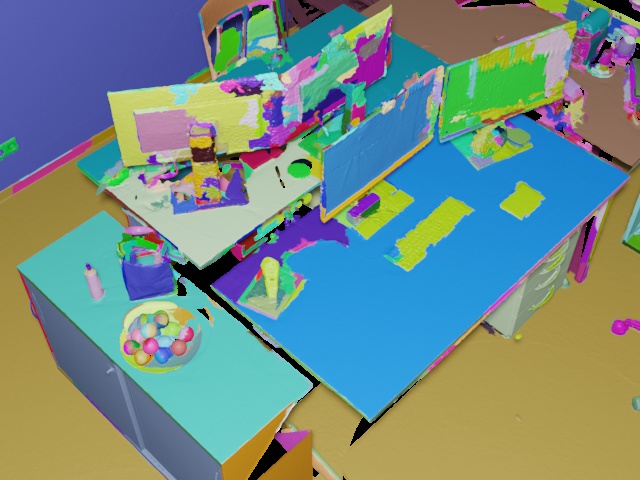}
    \hspace{-4mm}
    &\includegraphics[width=0.18\linewidth]{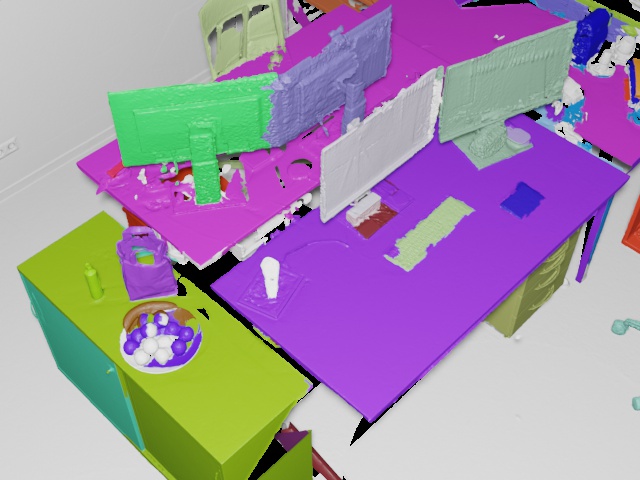}
    \hspace{-4mm}
    &\includegraphics[width=0.18\linewidth]{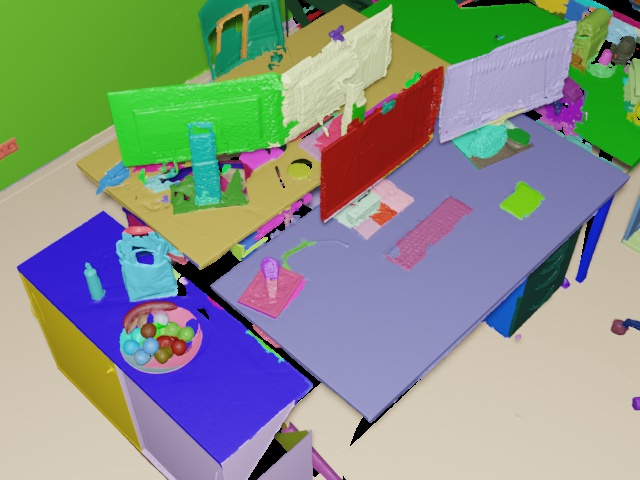}
    \hspace{-4mm}
    &\includegraphics[width=0.18\linewidth]{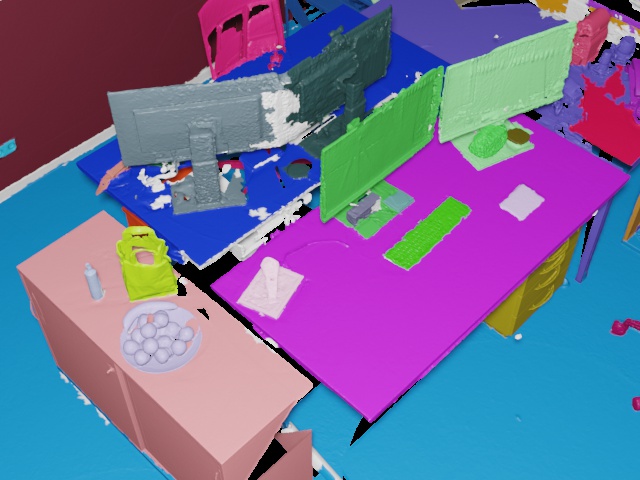} \\

    \includegraphics[width=0.18\linewidth]{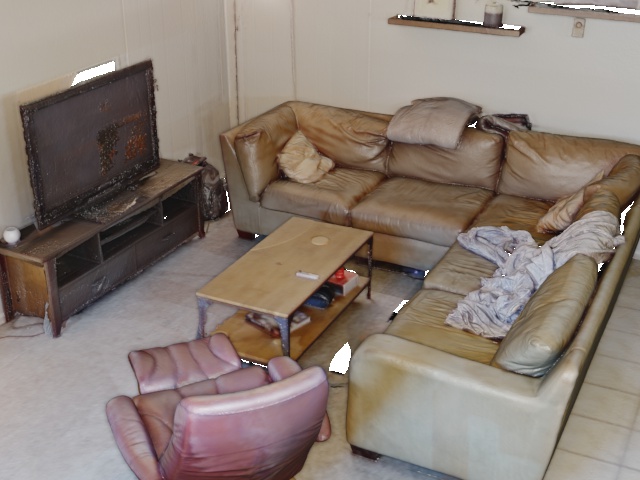}
    \hspace{-4mm}
    &\includegraphics[width=0.18\linewidth]{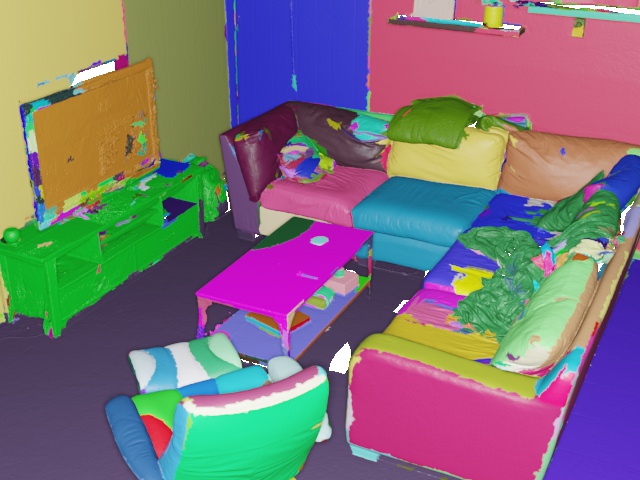}
    \hspace{-4mm}
    &\includegraphics[width=0.18\linewidth]{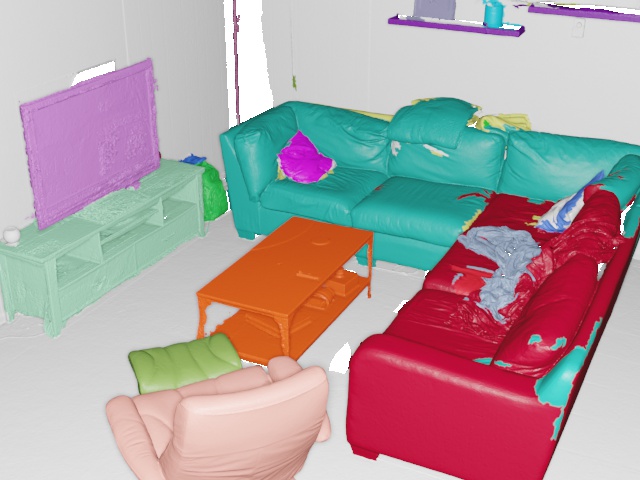}
    \hspace{-4mm}
    &\includegraphics[width=0.18\linewidth]{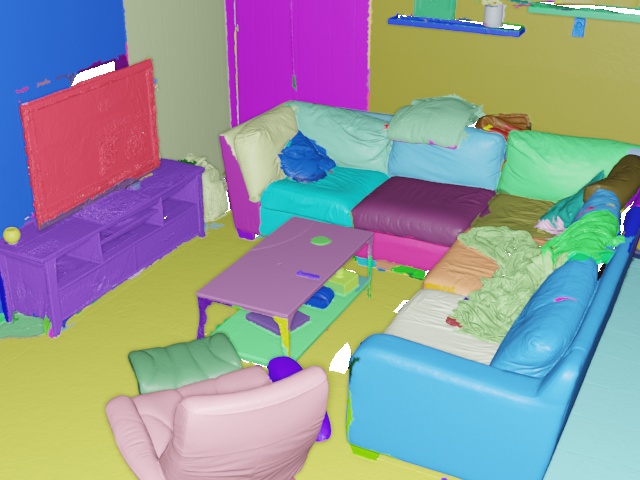}
    \hspace{-4mm}
    &\includegraphics[width=0.18\linewidth]{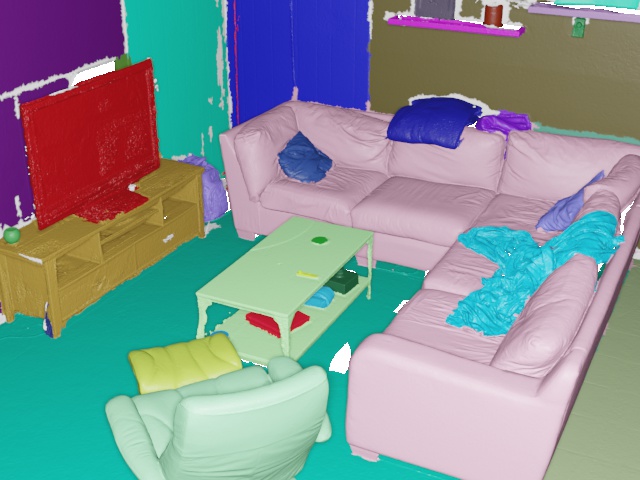} \\

    \includegraphics[width=0.18\linewidth]{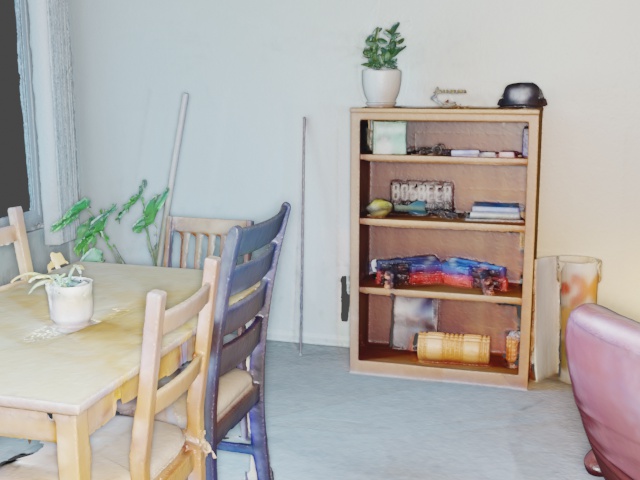}
    \hspace{-4mm}
    &\includegraphics[width=0.18\linewidth]{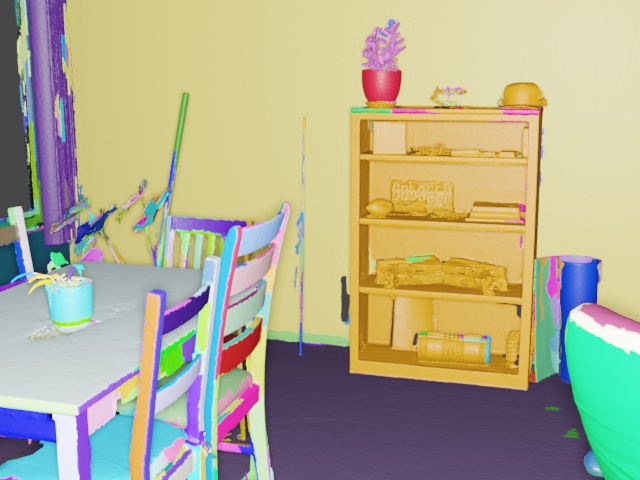}
    \hspace{-4mm}
    &\includegraphics[width=0.18\linewidth]{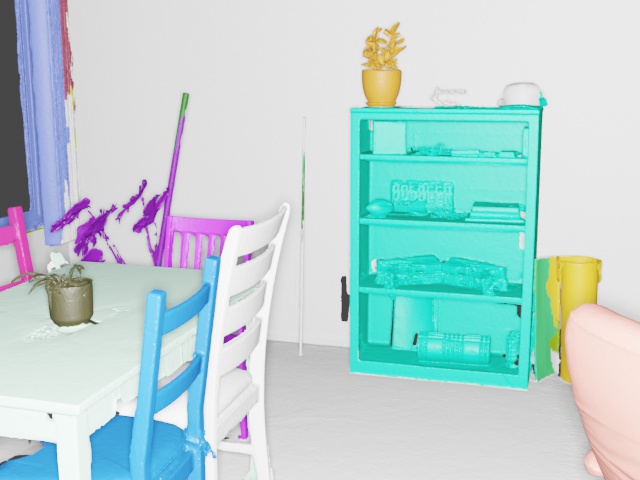}
    \hspace{-4mm}
    &\includegraphics[width=0.18\linewidth]{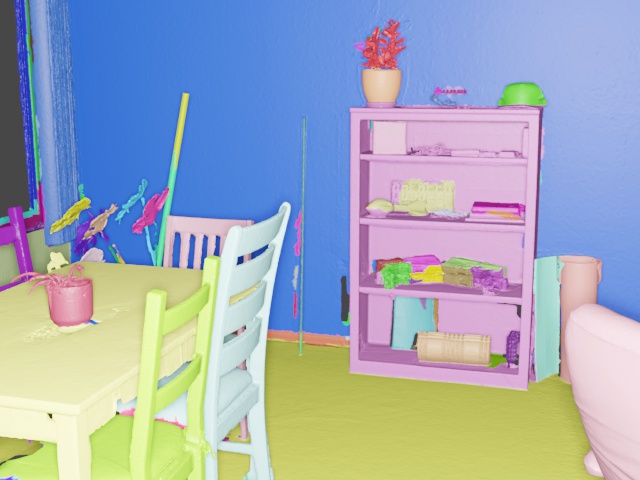}
    \hspace{-4mm}
    &\includegraphics[width=0.18\linewidth]{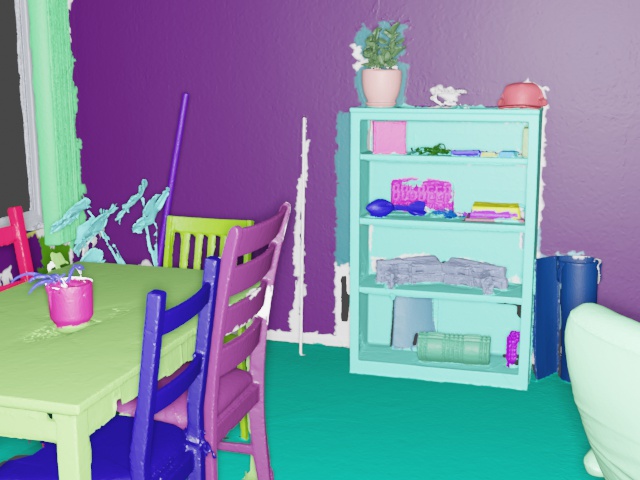} \\

    \small{Input} \hspace{-4mm}
    &\small{SAM3D} \hspace{-4mm} &\small{Mask3D} \hspace{-4mm} &\small{Ours} \hspace{-4mm} & \small{GT} 
    \end{tabular}
    \vspace{-2mm}
    \caption{\small {\textbf{Visual results of 3D instance segmentation on ScanNet++ dataset.} 
    We compare with both open-vocabulary and closed-vocabulary baselines.}
    }
	\label{fig:visual_scannetpp}
\end{figure*}

The visual comparisons, illustrated in Fig. \ref{fig:visual_scannetpp}, further underscore the effectiveness of our approach. 
Compared with prior works that struggle to generate clean segments on objects of small size, our approach is capable of identifying complex objects in cluttered scenes. For example, we succeed in segmenting various items stored in the cabinet while other methods group them together as a single instance.

\begin{figure*}[t]
    \centering
    \begin{tabular}{cccc}
    
    \vspace{-0.5mm}
    \includegraphics[width=0.22\linewidth]{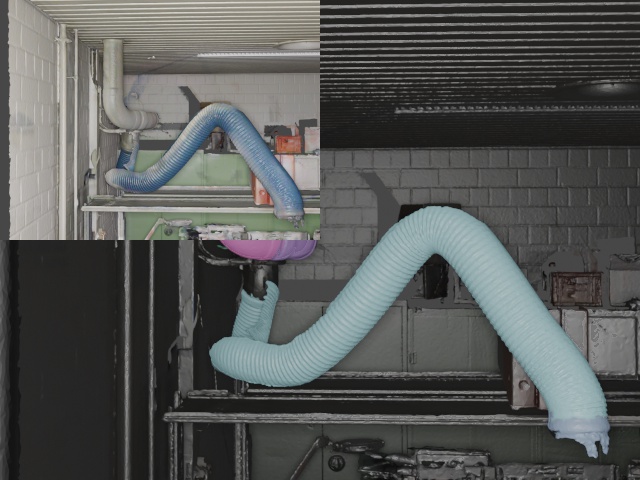}
    \hspace{-3mm}
    &\includegraphics[width=0.22\linewidth]{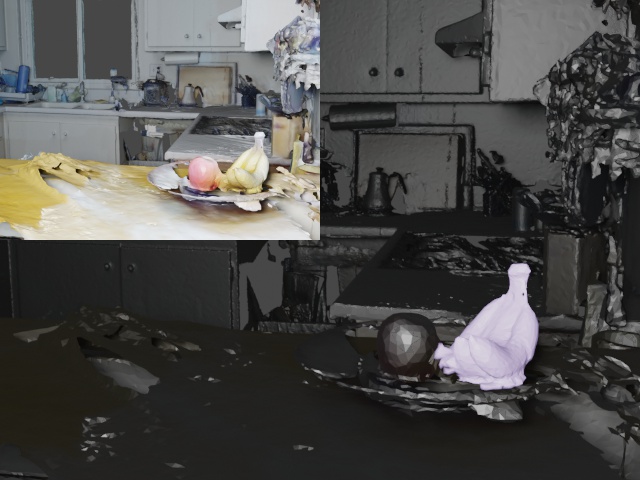}
    \hspace{-3mm}
    &\includegraphics[width=0.22\linewidth]{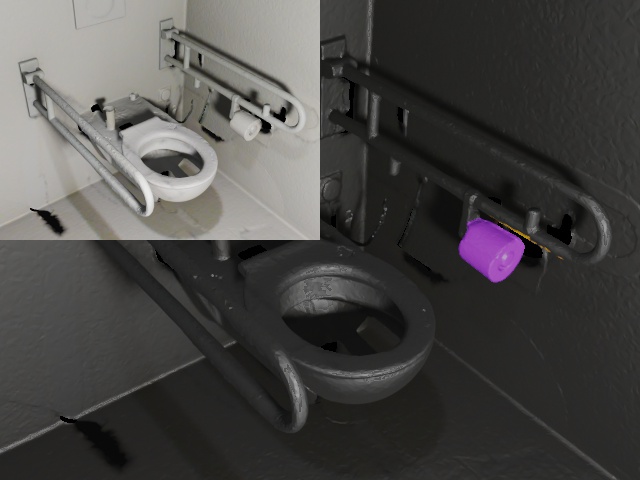} 
    \hspace{-3mm}
    &\includegraphics[width=0.22\linewidth]{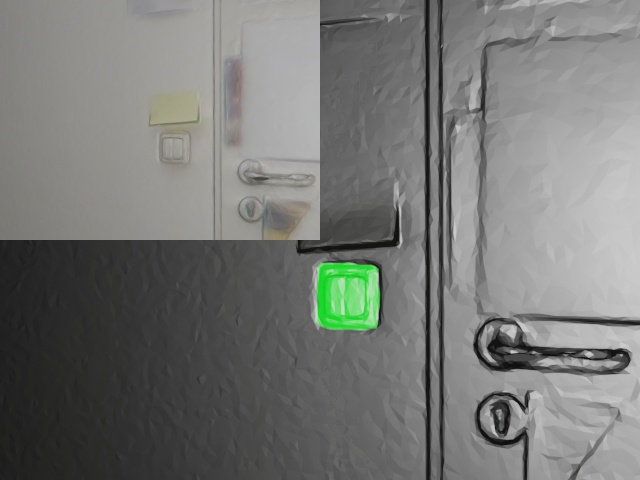} \\

    \vspace{1mm}
    \small{``Pipe''} &\small{``Banana''} &\small{``Toilet roll''} &\small{``Switch''} \\

    \includegraphics[width=0.22\linewidth]{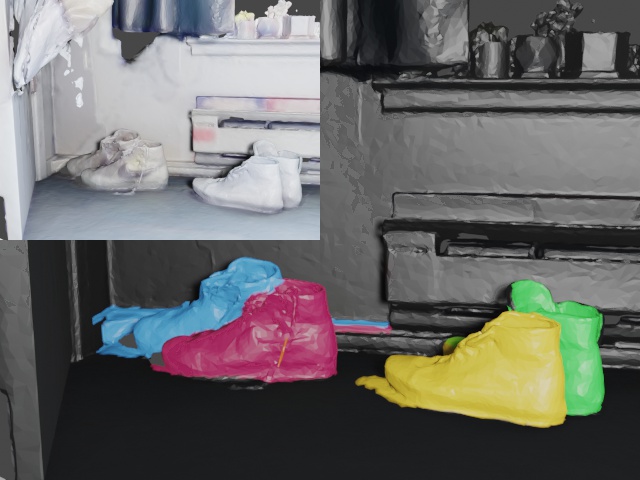}    
    \hspace{-3mm}
    &\includegraphics[width=0.22\linewidth]{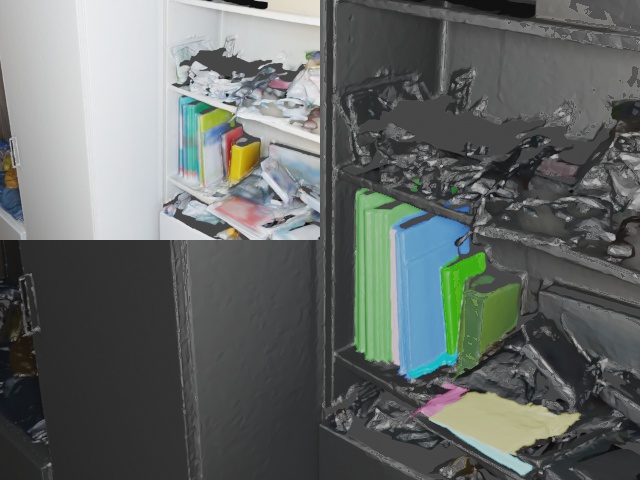}
    \hspace{-3mm}
    &\includegraphics[width=0.22\linewidth]{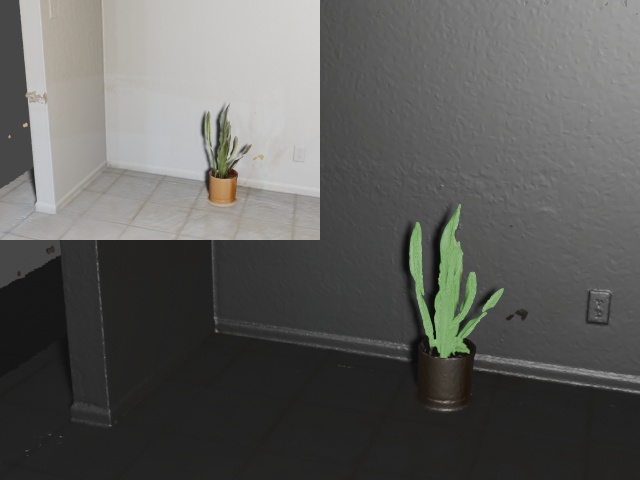}
    \hspace{-3mm}
    &\includegraphics[width=0.22\linewidth]{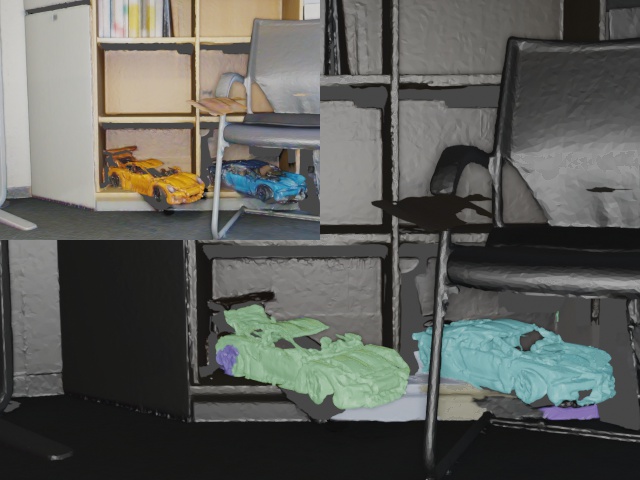} \\
    
    \small{``Shoes''} & \small{``Book''} &\small{``Botany''} &\small{``Toy car''}  

    \end{tabular}
    \vspace{-2mm}
    \caption{\small {\textbf{Open-vocabulary 3D object search.} Given a text prompt, our method finds accurate target object masks, even with long-tail semantics. 
    }}
    \vspace{-4mm}
	\label{fig:visual_scannetpp_semantic}
\end{figure*}

\vspace{1mm}
\noindent \textbf{Open-vocabulary 3D object querying.}
An important application of our method is zero-shot prompt-based instance segmentation. 
As depicted in Fig.~\ref{fig:teaser} and \ref{fig:visual_scannetpp_semantic}, our approach effectively segments the 3D scene into clean segments, and identifies specific objects with the input prompts.
Since our approach generates more detailed and accurate 3D instance segmentation than previous works, we are able to retrieve rare objects like ``toilet roll'' or ``shower gel''.

\begin{table}
    \centering
    \footnotesize
    \caption{\textbf{Class-agnostic 3D instance segmentation on ScanNetV2 dataset.} 
    }
    \vspace{-2mm}
    \begin{tabular}{lcccc}
    \toprule
        Method & Training Set & AP & AP$_{50}$ & AP$_{25}$  \\
        \midrule 
        \textit{Closed-vocabulary} \\
        Mask3D \cite{Schult23ICRA} & ScanNetV2 & 65.7 & 83.1 & 91.0 \\
        Mask3D \cite{Schult23ICRA} & S3DIS & 31.1 & 44.9 & 58.0 \\
        \midrule
        \textit{Open-vocabulary} \\
        HDBSCAN \cite{mcinnes2017accelerated} & - & 1.6 & 5.5 & 32.1 \\
        Nunes et al. \cite{nunes2022unsupervised} & - & 2.3 & 7.3 & 30.5 \\
        Felzenszwalb et al. \cite{felzenszwalb2004efficient} & - & 5.0 & 12.7 & 38.9 \\
        UnScene3D \cite{rozenberszki2023unscene3d} \footnote{Due to the unavailability of the code, we follow its experiment settings to evaluate on ScanNetV2 dataset.} & - & 15.9 & 32.2 & 58.5 \\
        SAM3D \cite{yang2023sam3d} & - & 20.2 & 34.0 & 53.3 \\
        Ours & - & \textbf{30.8} & \textbf{50.5} & \textbf{70.6} \\
    \bottomrule
    \end{tabular}
    \vspace{-2mm}
    \label{tab:class_agnotic_scannet}
\end{table}

\vspace{1mm}
\noindent \textbf{Standard 3D segmentation.}
Table~\ref{tab:class_agnotic_scannet} reports the class-agnostic instance segmentation results on ScanNetV2 dataset. Our method significantly outperforms the open-vocabulary baselines that do not require any training or finetuning on 3D annotation. 
As seen from the table, the supervised method Mask3D suffers from a performance drop when trained on a different dataset, which
indicates that supervised methods tend to overfit the training data, and lack a generalization ability toward open scenes.

\begin{table}[t]
  \centering
  \footnotesize
  \setlength{\tabcolsep}{0.5em}
  \caption{\textbf{Semantic instance segmentation on ScanNet200 dataset.}}
  \vspace{-2mm}
    \begin{tabular}{lcccccc}
    \toprule
        Method &  AP & AP$_{50}$ & AP$_{25}$ & \scriptsize \makecell{Head\\(AP)} & \scriptsize \makecell{Common\\(AP)} & \scriptsize \makecell{Tail\\(AP)} \\
        \midrule
        \multicolumn{7}{l}{\textit{Sup. mask + Open-vocab. semantic}} \\
        OpenMask3D \cite{takmaz2023openmask3d}   & {15.4} & {19.9} & 23.1 & {17.1} & {14.1} & {14.9}\\
        \midrule
        \multicolumn{7}{l}{\textit{Open-vocab. mask + Open-vocab. semantic}} \\
        OVIR-3D \cite{lu2023ovir}\footnote{The numbers are different from those in the original paper, since the original paper adopts a different evaluation metric namely \textit{mAP for information retrieval.}}         & 9.3 & 18.7 & \textbf{25.0} & 9.8 & 9.4 & 8.5 \\
        SAM3D \cite{yang2023sam3d}           & 9.8 & 15.2 & 20.7 & 9.2 & 8.3 & 12.3 \\
        Ours            & \textbf{12.7}  & \textbf{18.8} & {24.1} & \textbf{12.1} & \textbf{10.4} & \textbf{16.2} \\
    \bottomrule
    \end{tabular}
    \vspace{-2mm}
    \label{tab:semantic_instance_scannet}
\end{table}

\yd{Following OpenMask3D \cite{takmaz2023openmask3d}, \textit{semantic} instance segmentation is evaluated on ScanNet200 dataset}, with numerical results in Table \ref{tab:semantic_instance_scannet}.
\yd{We adopt the pipeline of OpenMask3D \cite{takmaz2023openmask3d} to assign semantic labels for our method.}
As shown in the table, our method clearly results in superior performance than the open-vocabulary baselines, but lags behind \yd{OpenMask3D that relies on the supervised mask proposals}.
Upon closer examination, we notice that our method shows weaker performance on frequently occurring semantics (Head), but outperforms on long-tail labels (Tail). This observation highlights the strength of our method in zero-shot generalization, better at handling diverse and less common labels.

\yd{
To further evaluate the generalizability, we conduct experiments on Matterport3D dataset for both class-agnostic and semantic instance segmentation tasks. As shown in Table \ref{tab:matterport}, our zero-shot method consistently outperforms other baselines, while OpenMask3D suffers a significant performance drop since it is trained on ScanNet200 dataset.}

\begin{table}[t]
  \centering
  \footnotesize
  \setlength{\tabcolsep}{0.65em}
    \caption{\textbf{Class-agnostic and semantic instance segmentation on Matterport3D dataset.}}
    \vspace{-2mm}
    \begin{tabular}{lcccccc}
    \toprule
        \multirow{2}{*}{Method} & \multicolumn{3}{c}{Class-agnostic} & \multicolumn{3}{c}{Semantic} \\
        &  AP & AP$_{50}$ & AP$_{25}$  &  AP & AP$_{50}$ & AP$_{25}$ \\
        \midrule
        \footnotesize OpenMask3D \cite{takmaz2023openmask3d} &  15.3 & 28.3  &  43.3  & 7.7  & 13.9  & 20.3 \\
        \footnotesize OVIR-3D \cite{lu2023ovir} & 6.6  &  15.6 & 28.3  & 5.8  & 13.8  & \textbf{22.4} \\
        \footnotesize SAM3D \cite{yang2023sam3d} & 10.1  &  19.4 &  36.1 &  4.4 & 7.3  & 11.3 \\
        \footnotesize Ours &  \textbf{21.5} & \textbf{38.3}  & \textbf{59.1}  & \textbf{8.9}  & \textbf{15.3}  & 20.9 \\
    \bottomrule
    \end{tabular}
    \label{tab:matterport}
\end{table}

\subsection{Ablation and Analysis}

\setlength{\tabcolsep}{4pt}
\begin{table}
    \centering
    \footnotesize
    \caption{\textbf{The ablation studies of the effectiveness of our designs.} The experiments are conducted on ScanNetV2 dataset.}
    \vspace{-2mm}
    \begin{tabular}{cccccc}
    \toprule
        \makecell[c]{3D superpoint \\ primitive} &  
        \makecell[c]{Multi-level \\ merging criteria} &  
        \makecell[c]{Progressive \\ growing}  &  
        AP  & 
        AP$_{50}$  & 
        AP$_{25}$ \\
        \midrule 
         &  &   &  19.5 &  32.8 & 52.9 \\
         
        \ding{51}  &   &  &  24.2 &  39.5 & 59.2 \\
        
        \ding{51}  & \ding{51}  &  & 28.4 & 46.5 & 67.0 \\
        
        \ding{51}  & & \ding{51}  & 27.6 &  45.5 & 65.8 \\
        
        \ding{51}  & \ding{51}  & \ding{51}   & \textbf{30.8}  &  \textbf{50.5} &  \textbf{70.6} \\
        
    \bottomrule
    \end{tabular}
    \vspace{-2mm}
    \label{tab:ablation_method}
\end{table}

\noindent \textbf{Effect of our designs on the region growing algorithm.}
We analyze the three key design choices of our region growing algorithm: the adoption of \textit{3D superpoint primitive}, \textit{Multi-level merging criteria}, and \textit{Progressive growing}. The results of ablation studies are shown in Table \ref{tab:ablation_method}.
We find that omitting the 3D superpoints and using a classic point-based region-growing approach leads to a noticeable drop in performance. The incorporation of multi-level merging criteria and the use of progressive growing both contribute to enhancing the algorithm’s robustness, particularly in dealing with the noise present in 2D segmentation masks, yielding better results respectively.
When all the proposed designs are combined, there is an approximately 50\% improvement in performance compared to the original, basic version of the algorithm, underlining the effectiveness of our design choices of the algorithm.

\begin{figure}[!t]
\centering
    \includegraphics[width=1.\linewidth]{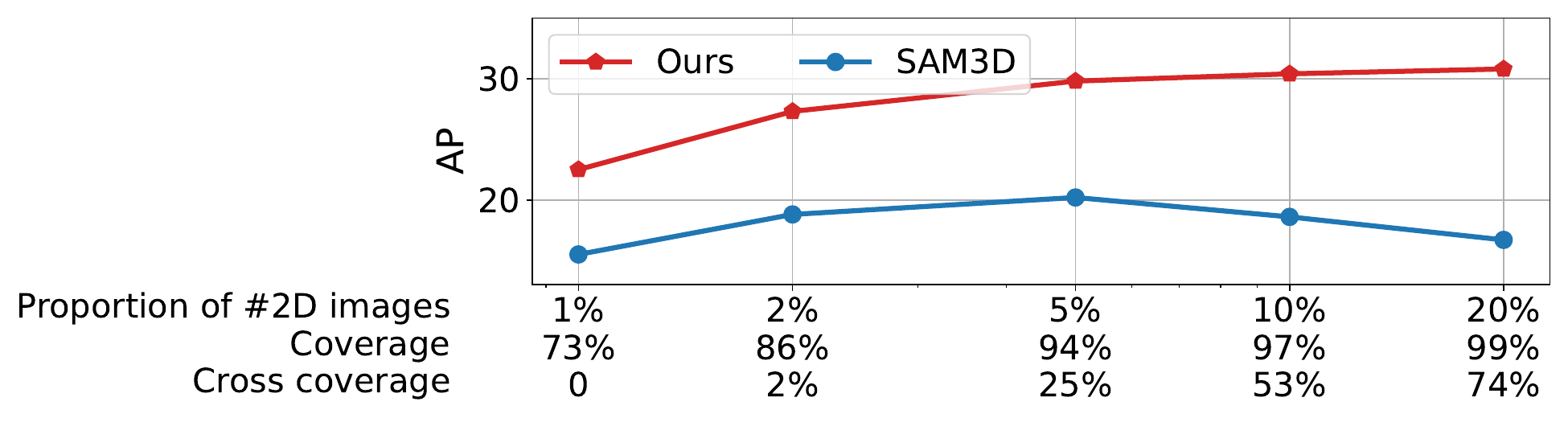}
    \vspace{-4mm}
    \caption{
    \textbf{Ablation studies on the correlation between performance and the number of 2D images.} 
    \yd{We define \textit{coverage} and \textit{cross coverage} as the percentage of points observed at least once and more than 10 times, respectively. The experiments are conducted on ScanNetV2 dataset.}
    }
 \vspace{-4mm}
    \label{fig:view_proportion}
\end{figure}

\vspace{1mm}
\noindent \textbf{Effect of the number of 2D images.}
As both our approach and the concurrent method SAM3D leverages 2D image masks, we thus study the method's robustness against the number of images.
Among all the images of a 3D scene, we vary the proportion of views from 1\% to 20\% and report the performance curve in Fig.~\ref{fig:view_proportion}.
Our approach progressively improves the 3D segmentation performance when more views are provided, while SAM3D fails to aggregate useful information when the proportion of views becomes larger than 5\%.
More importantly, our approach achieves an AP score larger than 20 with only 1\% of views, outperforming the best performance achieved by SAM3D.
This highlights the effectiveness of our 3D primitive-based region-growing algorithm.

\section{Conclusion}\label{sec:conclusion}

We introduce \ours, a novel approach for zero-shot 3D instance segmentation. Our method offers an efficient alternative to supervised methods which are confined to specific datasets and hence inhibit broader applicability in open-world scenarios. 
The technique we presented builds upon the power of 2D and 3D segmentation techniques, and a novel progressive region-growing algorithm, which smartly merges the 3D superpoints into object masks that define the final 3D instance segmentation. 

Through our evaluation on ScanNet, ScanNet++ and Matterport3D datasets, we demonstrate that our approach significantly outperforms prior unsupervised methods like SAM3D in class-agnostic segmentation,
and even surpass fully supervised Mask3D models trained with 3D annotations on the more challenging ScanNet++ dataset. 
This success underscores the potential of leveraging geometric primitives and multi-view consistency to achieve high-quality instance segmentation in complex 3D scenes, without using 3D labeled data.

\vspace{1mm}
\noindent \textbf{Limitations.}
Our method fundamentally relies on accurate 2D segmentation results \yd{and reliable 2D-3D alignment}. 
As our approach builds a scene graph based on 3D primitives and computes an affinity matrix based on 2D masks \yd{through 2D-3D lifting}, incorrect 2D masks \yd{or camera poses} naturally result in unreliable affinities, thus affecting the fine segmentation results.
\yd{Though several techniques are designed to better leverage geometry priors and multi-view consensus,}
designing a more advanced 2D mask aggregation mechanism should be a promising direction.

Another limitation is the running speed. 
As our approach involves aggregating 2D mask segmentation from images, the total processing time linearly scale with the number of images making it difficult to apply in large-scale scenes.
Designing a more efficient algorithm without iterating over all images is another important future direction.

{
    \small
    \bibliographystyle{ieeenat_fullname}
    \bibliography{main}
}
\appendix

\section{Implementation Details}

\paragraph{Multi-level merging criteria.}
To compute the affinity score $A_{R,Q_i}$ between a region $R$ and a superpoint $Q_i$, we consider all the affinity scores $A_{i,k}$ between $Q_i$ and the superpoints inside the region $\{Q_k\}\in R$. Specifically, it is computed as the weighted average 
\begin{equation}
    A_{R,Q_i} = \frac{1}{\sum_{Q_k \in R}\beta_{i,k}} \sum_{Q_k \in R} \beta_{i,k} ~ A_{i,k}
\end{equation}
where $A_{i,k}$ is the affinity between $Q_i$ and $Q_k$, and $\beta_{i,k}$ is the weight factor indicating how much $A_{i,k}$ contribute to $A_{R,Q_i}$. $\beta_{i,k}$ is determined by the distance $d_{i,k}$ between $Q_i$ and $Q_k$, as well as the number of points $N_k$ inside $Q_k$.
\begin{equation}
    \beta_{i,k} = 
        \left\{ 
            \begin{array}{lc}
                \gamma^d N_k & d \leq 2 \\
                0 &  d > 2\\
            \end{array}
        \right.
\end{equation}
For superpoints that can directly reach $Q_i$, we define the distance as $d=1$. Similarly, the set of superpoints that reach $Q_i$ through one bridging superpoint is with $d=2$, and so forth. We eliminate the contributions when $d > 2$ for computational efficiency. $\gamma$ is set as 0.5.

\paragraph{Progressive region growing.}
The {region-growing} algorithm is illustrated in Algo. \ref{algo:growing}.
For progressive growing, we build a multi-stage region-growing framework with the affinity threshold varying from high to low. 
Our algorithm benefits from the dynamic strategy and leads to robustness to the choice of the thresholds. We set [0.9, 0.8, 0.7] for ScanNet++ dataset and [0.9, 0.8, 0.7, 0.6, 0.5] for ScanNet dataset. We did not tune the hyper-parameters much.

\paragraph{Datasets.}
\yd{To balance the performance and the efficiency, we sample proportions of images for different datasets. We use $5\%$ views for ScanNet++ dataset, $20\%$ views for both ScanNetV2 and ScanNet200 dataset, and all images in Matterport3D since it is a sparsely-scanned dataset.}

\yd{Following the common practice \cite{schult2023mask3d,takmaz2023openmask3d,lu2023ovir}, we ignore the instances with semantics of ``wall'' and ``floor'' for ScanNetV2 and ScanNet200 datasets. For Matterport3D dataset, we consider the most frequent 160 classes provided in \cite{Peng2023OpenScene} and ignore ``wall'', ``floor'' and ``ceiling''. 
For ScanNet++ dataset, we use all GT labels provided for the instance segmentation task (in {``instance\_classes.txt''}).
IPhone images are served as our input to better reflect everyday cases.}

\paragraph{Evaluation.}
Following the baselines \cite{rozenberszki2023unscene3d,schult2023mask3d,takmaz2023openmask3d,Peng2023OpenScene}, we evaluate the numerical results on the validation set \yd{for ScanNetV2, ScanNet200 and ScanNet++ datasets and on the test set for Matterport3D dataset.}
We follow UnScene3D \cite{rozenberszki2023unscene3d} to implement \textit{class-agnostic} instance segmentation, where all object categories are treated equally and only the mask AP values are considered. 
\yd{We set all the confidence scores as 1.0, the same as \cite{takmaz2023openmask3d}.}
For 2D foundation models, we choose SAM-HQ \cite{sam_hq} for ScanNet++ dataset, and Semantic-SAM \cite{li2023semanticSAM} for \yd{ScanNetV2, ScanNet200 and Matterport3D} datasets for better granularity control.

\begin{algorithm}
    \small
    \caption{Region-growing algorithm}
    \label{algo:growing}
    
    \begin{algorithmic}[0] 
        \Statex \textbf{Input: } Affinity matrix $\textbf{A} \in \mathbb{R}^{N_Q \times N_Q}$ where $A_{i,j}$ indicates the affinity score between two super points $Q_i$ and $Q_j$, and $N_Q$ is the number of superpoints.
        \Statex \textbf{Output: } Instance label $\mathbf{l}\in\mathbb{R}^{N_Q}$.  

        \State $\mathbf{l}  \leftarrow \mathbf{0}$
        \State id $\leftarrow 1$
        \For {$i \leftarrow 1$ to $N_Q$}
            \If {$l_i = 0$}
                \State Queue $B$
                \State $B$.push($i$)
                \State $l_i \leftarrow $ id
                \While {$B$ not empty}
                    \State $v \leftarrow B$.pop()
                    \For {$j \leftarrow $ neighbors of $v$}
                        \If {$l_j \neq 0$}
                            \State continue
                        \EndIf
                        \State $R \leftarrow \{Q_k | l_k=$ id$\}$
                        \State $A_{R,Q_j} \leftarrow$ Multi-level\_criteria$(R, Q_j, \mathbf{A})$
                        \If {$A_{R,Q_j}  > \tau$}
                            \State $B$.push($j$)
                            \State $l_j \leftarrow$ id
                        \EndIf
                    \EndFor
                \EndWhile
                \State id $\leftarrow$ id $+1$
            \EndIf
        \EndFor
        
    \end{algorithmic}
\end{algorithm}

\section{Additional Qualitative Results}
\begin{figure*}[t]
    \centering
    \begin{tabular}{ccccc}

    \includegraphics[width=0.19\linewidth]{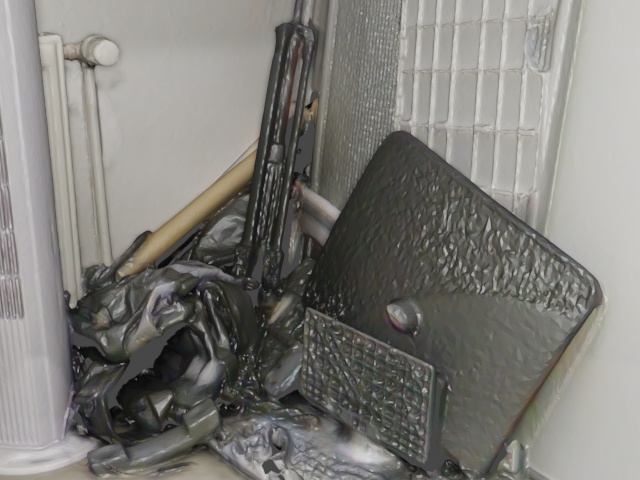}
    \hspace{-4mm}
    &\includegraphics[width=0.19\linewidth]{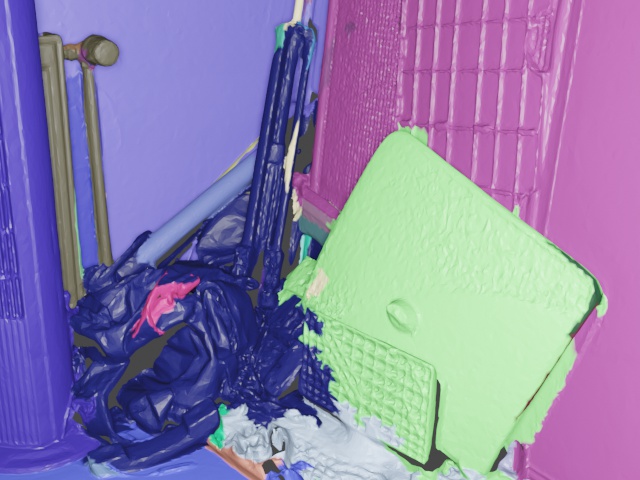}
    \hspace{-4mm}
    &\includegraphics[width=0.19\linewidth]{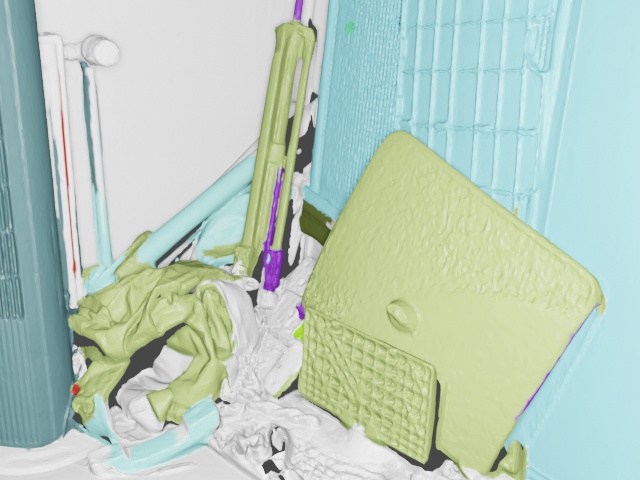}
    \hspace{-4mm}
    &\includegraphics[width=0.19\linewidth]{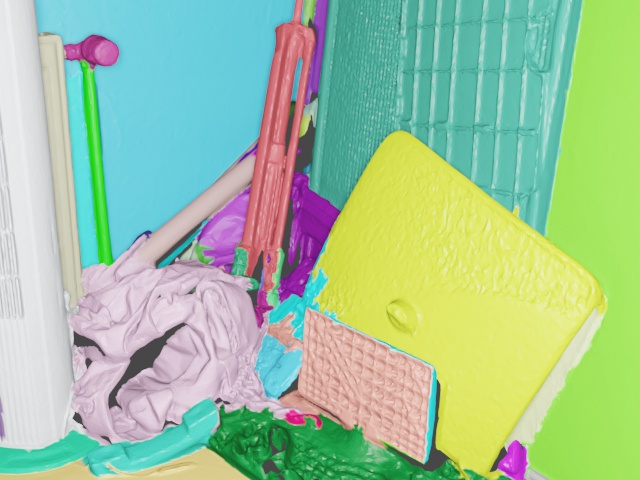}
    \hspace{-4mm}
    &\includegraphics[width=0.19\linewidth]{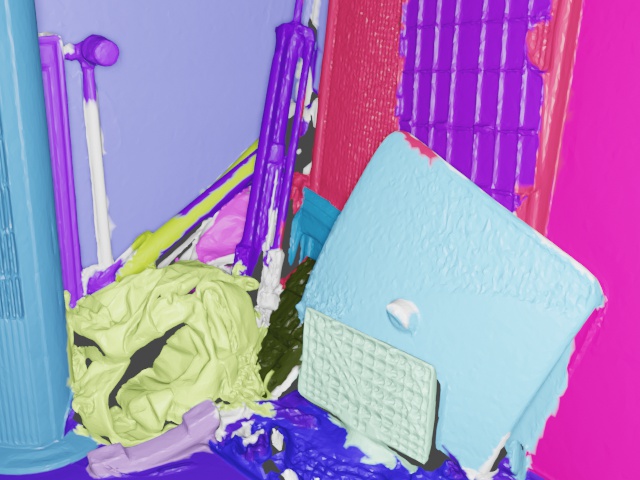} \\

    \includegraphics[width=0.19\linewidth]{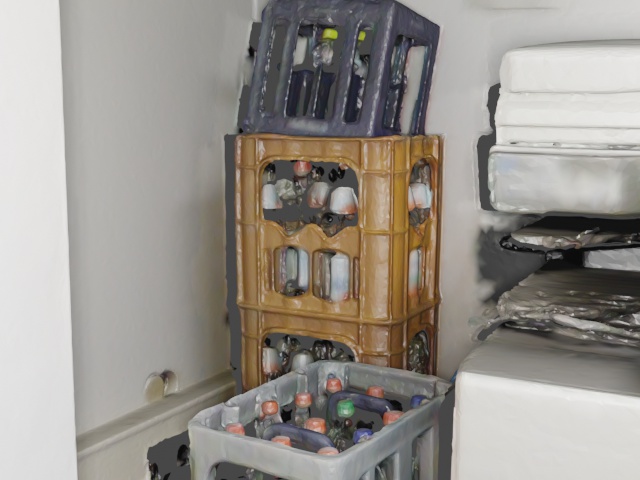}
    \hspace{-4mm}
    &\includegraphics[width=0.19\linewidth]{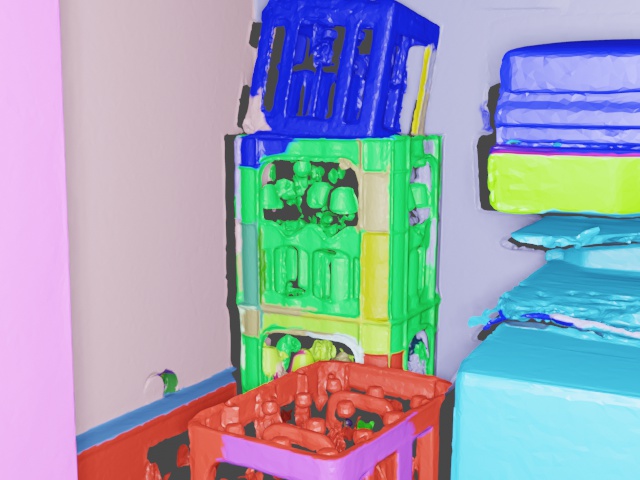}
    \hspace{-4mm}
    &\includegraphics[width=0.19\linewidth]{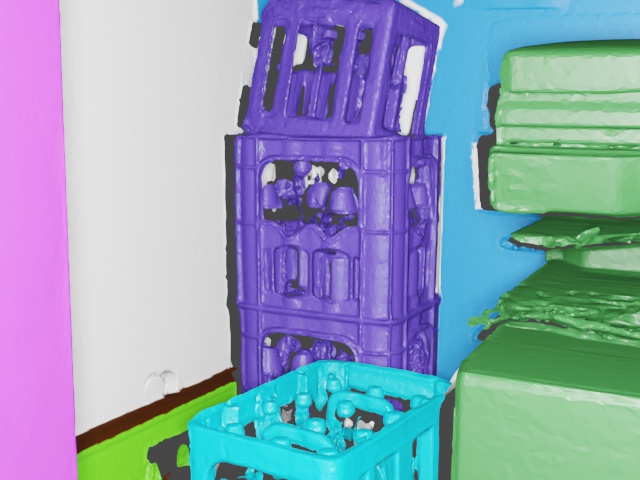}
    \hspace{-4mm}
    &\includegraphics[width=0.19\linewidth]{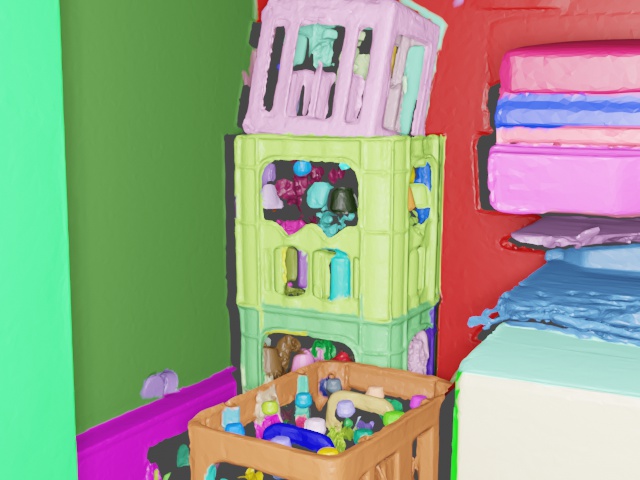}
    \hspace{-4mm}
    &\includegraphics[width=0.19\linewidth]{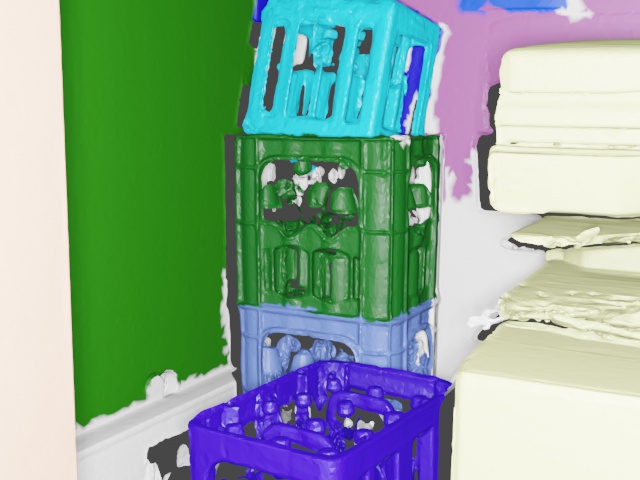} \\

    \includegraphics[width=0.19\linewidth]{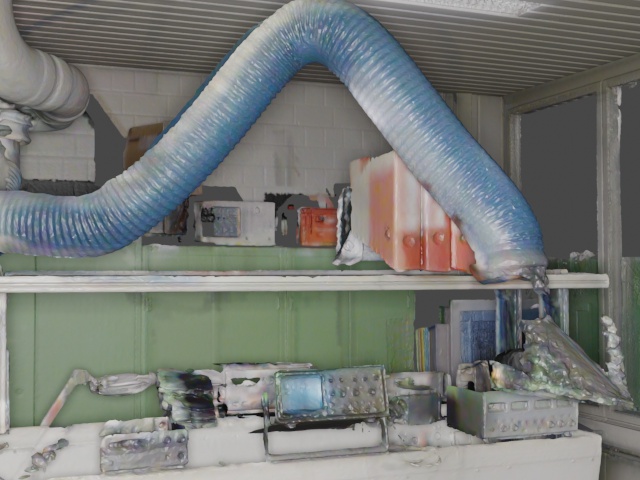}
    \hspace{-4mm}
    &\includegraphics[width=0.19\linewidth]{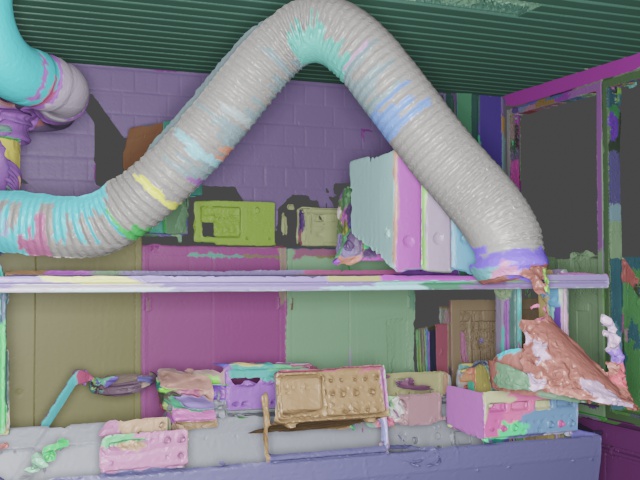}
    \hspace{-4mm}
    &\includegraphics[width=0.19\linewidth]{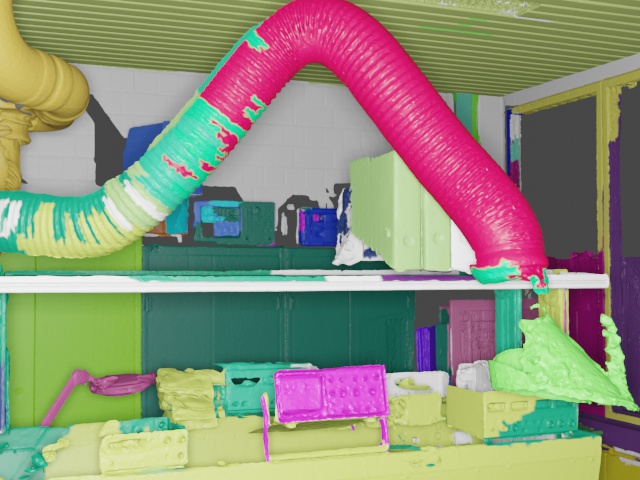}
    \hspace{-4mm}
    &\includegraphics[width=0.19\linewidth]{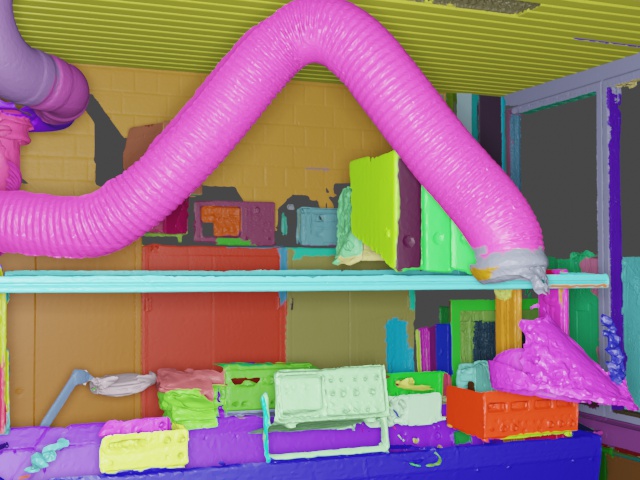}
    \hspace{-4mm}
    &\includegraphics[width=0.19\linewidth]{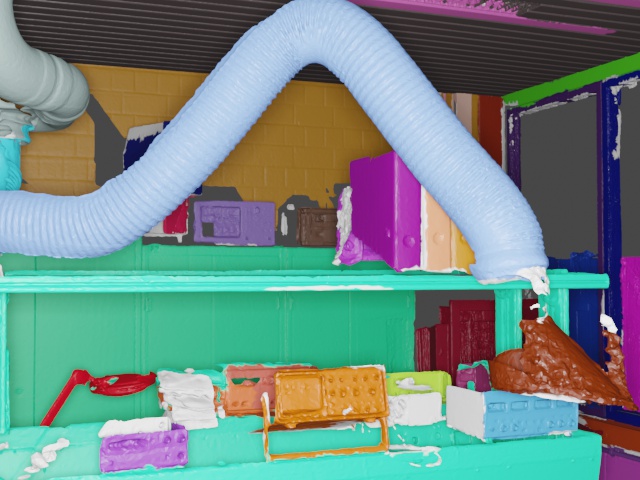} \\

    \includegraphics[width=0.19\linewidth]{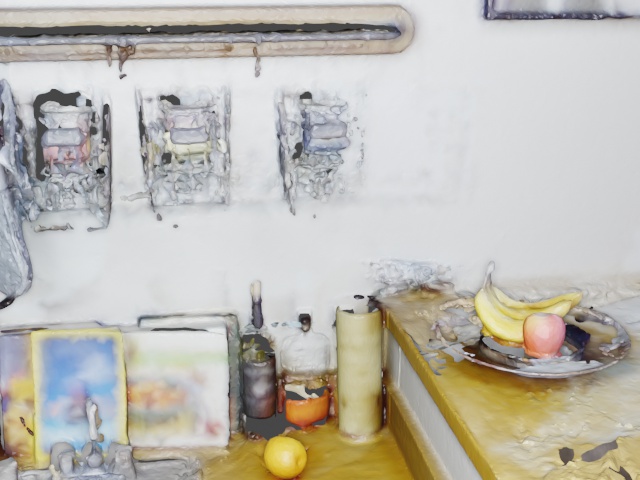}
    \hspace{-4mm}
    &\includegraphics[width=0.19\linewidth]{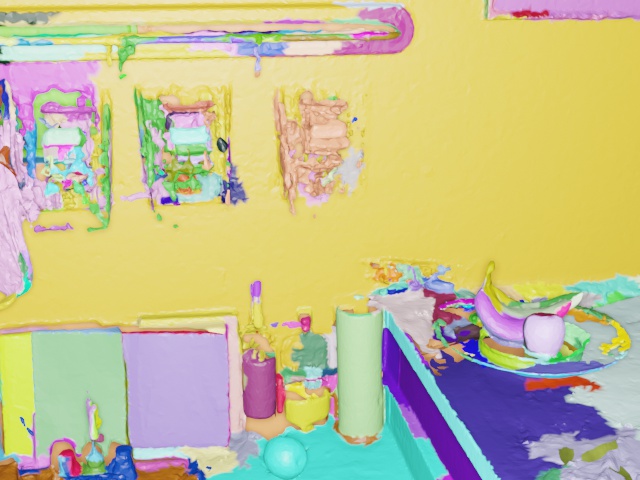}
    \hspace{-4mm}
    &\includegraphics[width=0.19\linewidth]{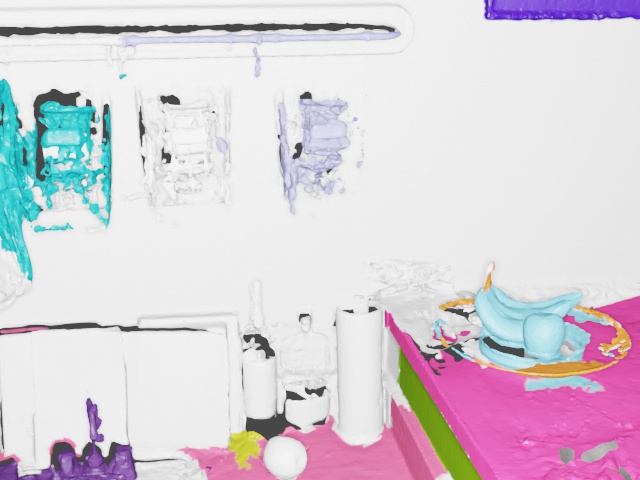}
    \hspace{-4mm}
    &\includegraphics[width=0.19\linewidth]{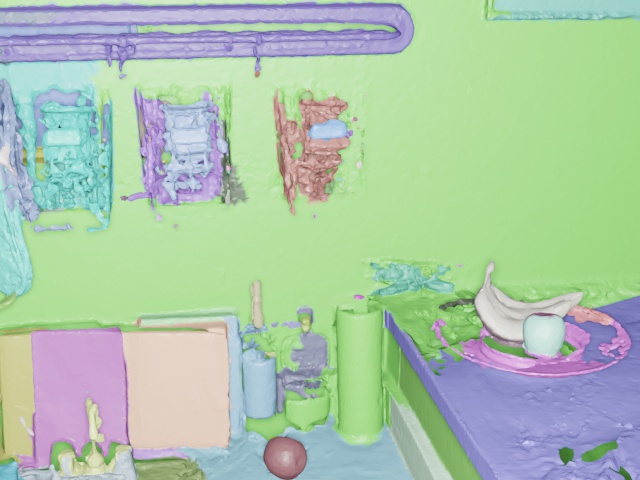}
    \hspace{-4mm}
    &\includegraphics[width=0.19\linewidth]{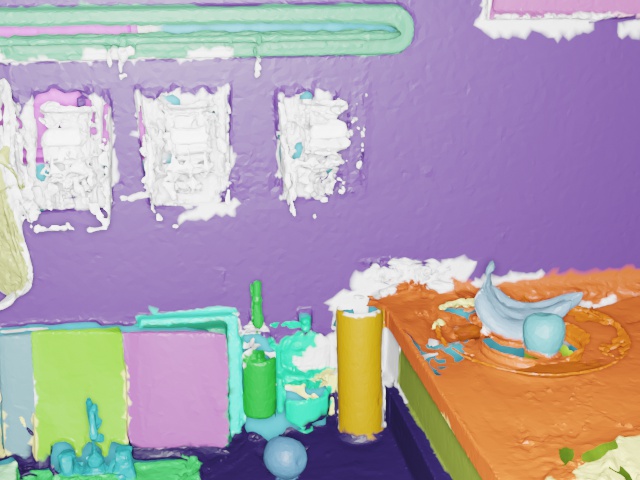} \\

    \includegraphics[width=0.19\linewidth]  {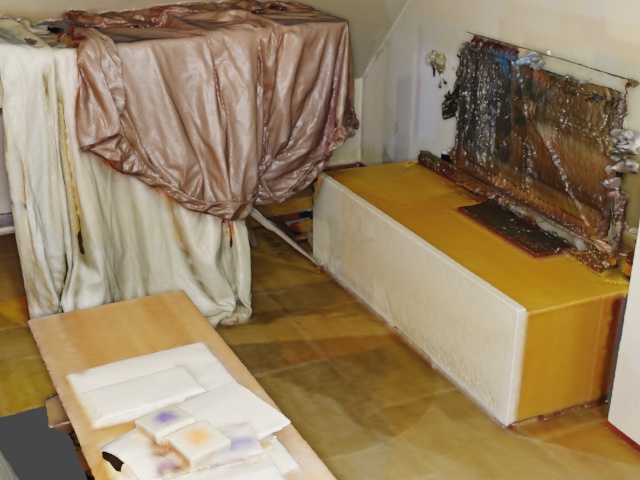} 
    \hspace{-4mm}
    &\includegraphics[width=0.19\linewidth]{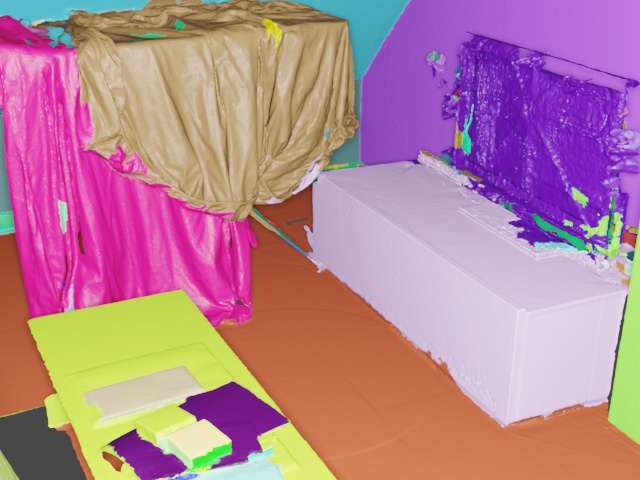}
    \hspace{-4mm}
    &\includegraphics[width=0.19\linewidth]{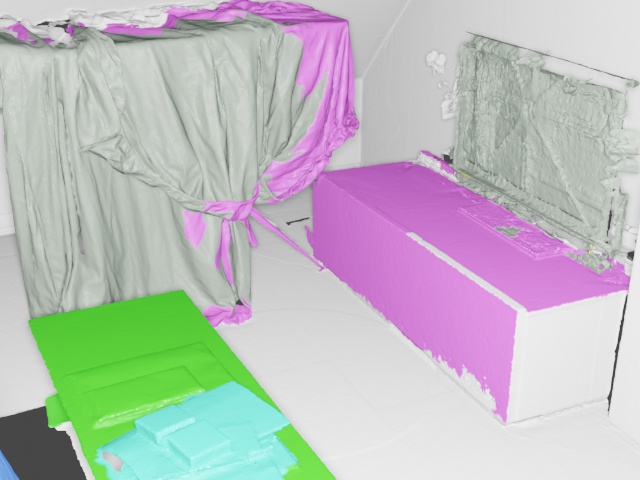}
    \hspace{-4mm}
    &\includegraphics[width=0.19\linewidth]{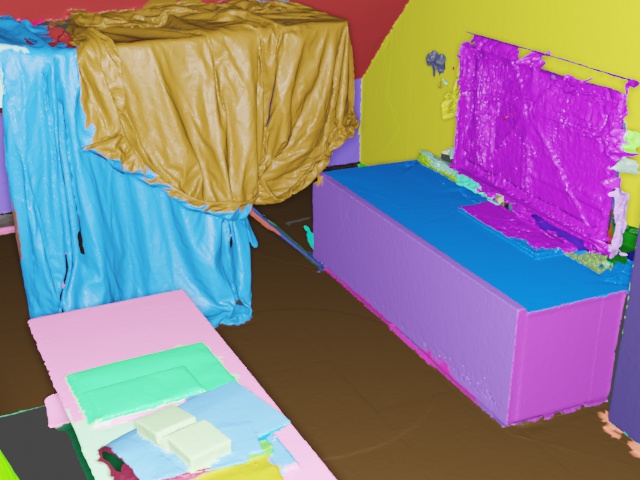}
    \hspace{-4mm}
    &\includegraphics[width=0.19\linewidth]{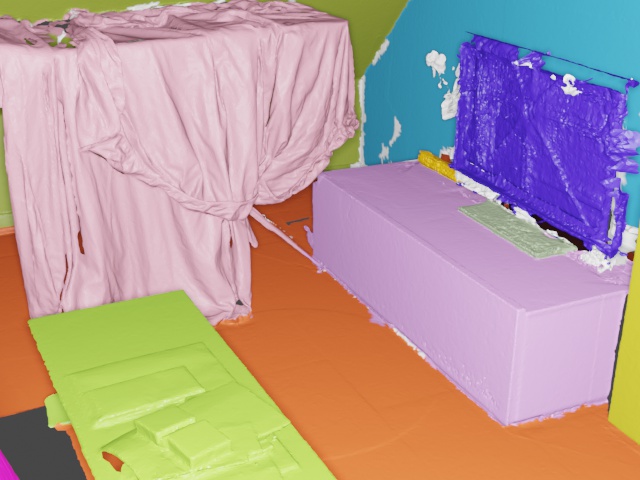} \\ 

    \includegraphics[width=0.19\linewidth]{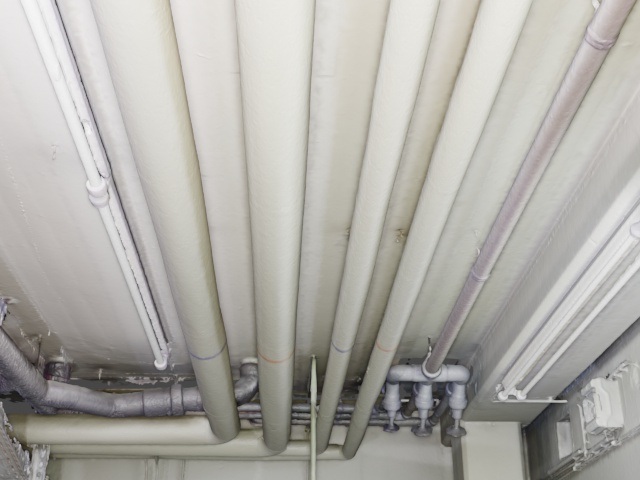} 
    \hspace{-4mm}
    &\includegraphics[width=0.19\linewidth]{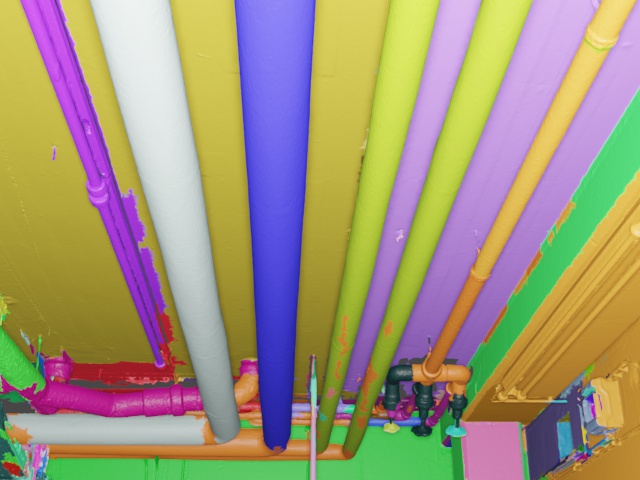}
    \hspace{-4mm}
    &\includegraphics[width=0.19\linewidth]{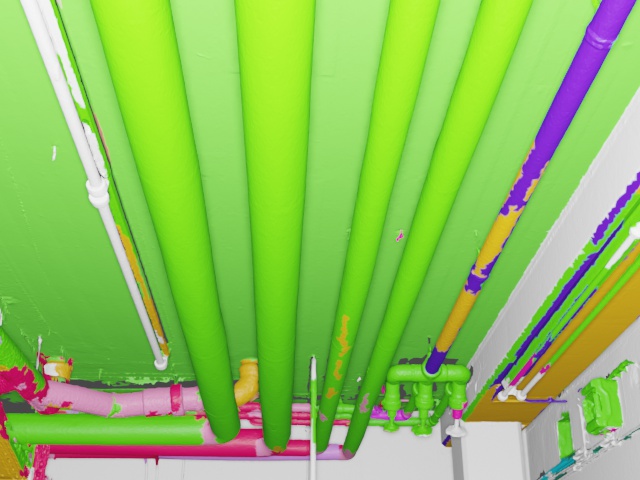}
    \hspace{-4mm}
    &\includegraphics[width=0.19\linewidth]{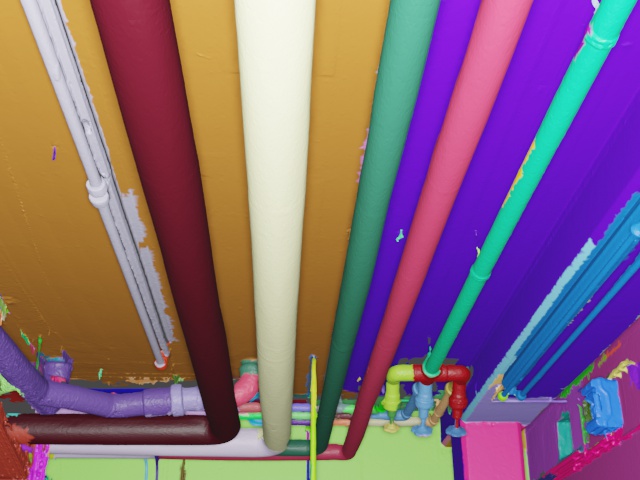}
    \hspace{-4mm}
    &\includegraphics[width=0.19\linewidth]{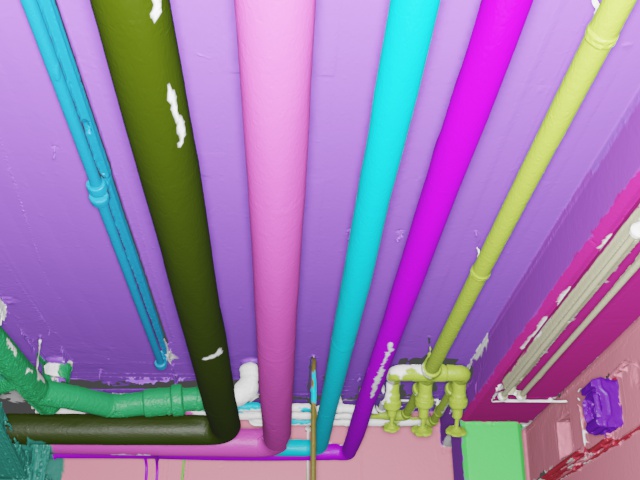} \\

    \includegraphics[width=0.19\linewidth]{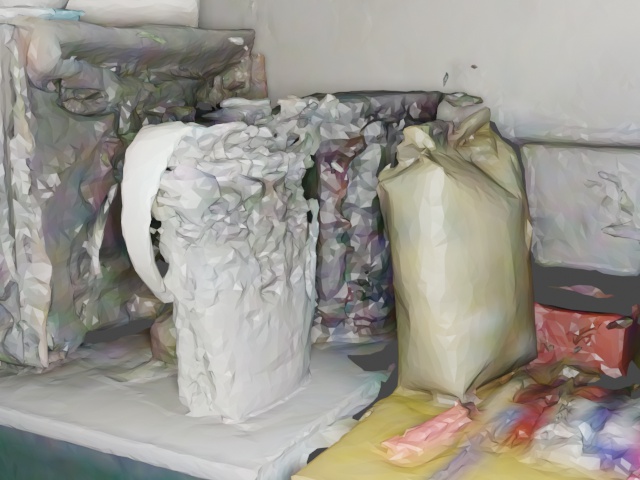}
    \hspace{-4mm}
    &\includegraphics[width=0.19\linewidth]{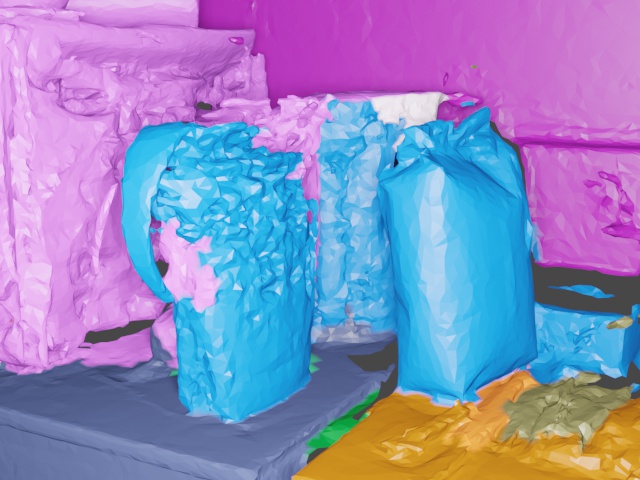}
    \hspace{-4mm}
    &\includegraphics[width=0.19\linewidth]{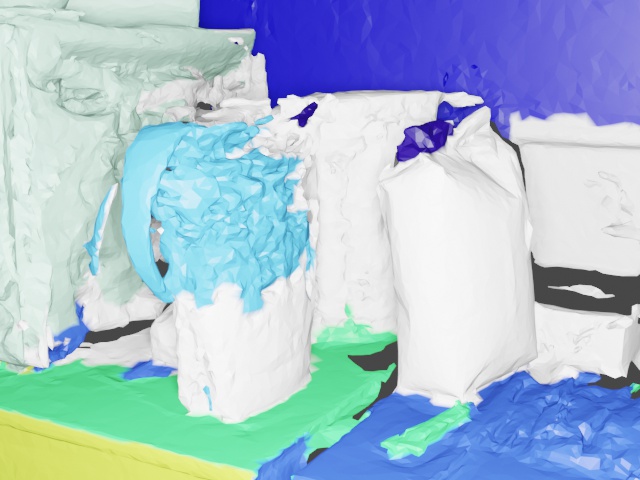}
    \hspace{-4mm}
    &\includegraphics[width=0.19\linewidth]{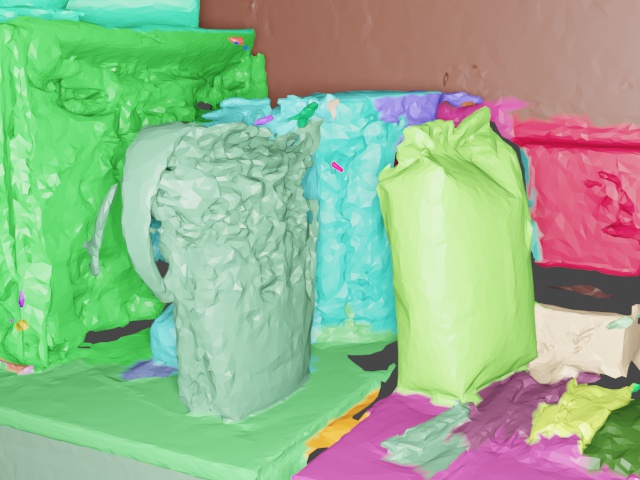}
    \hspace{-4mm}
    &\includegraphics[width=0.19\linewidth]{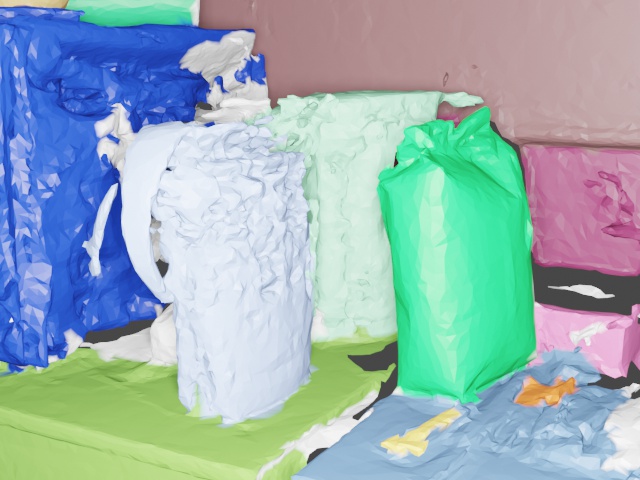} \\

    \includegraphics[width=0.19\linewidth]{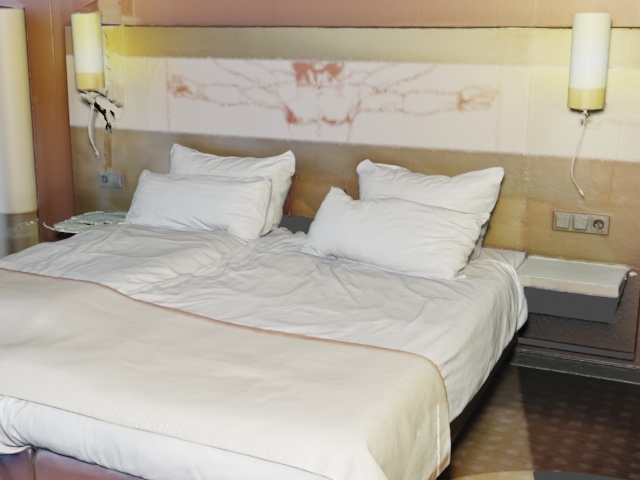}
    \hspace{-4mm}
    &\includegraphics[width=0.19\linewidth]{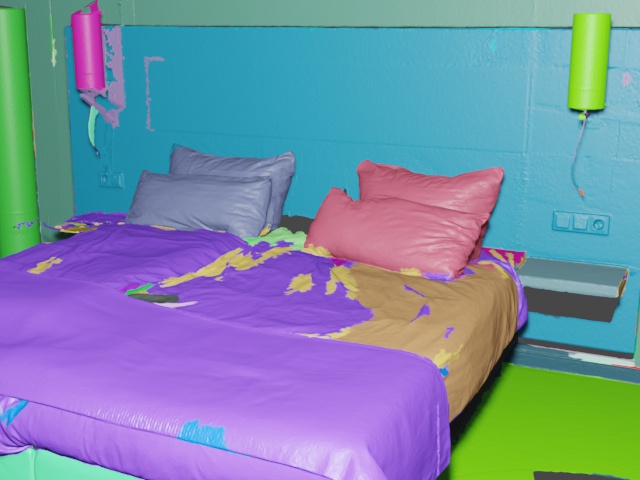}
    \hspace{-4mm}
    &\includegraphics[width=0.19\linewidth]{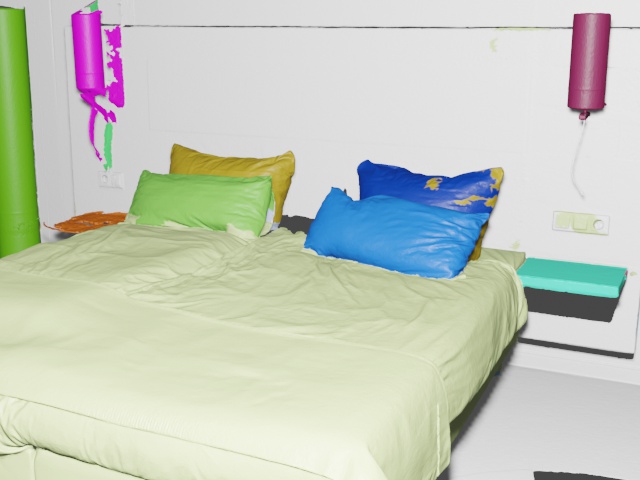}
    \hspace{-4mm}
    &\includegraphics[width=0.19\linewidth]{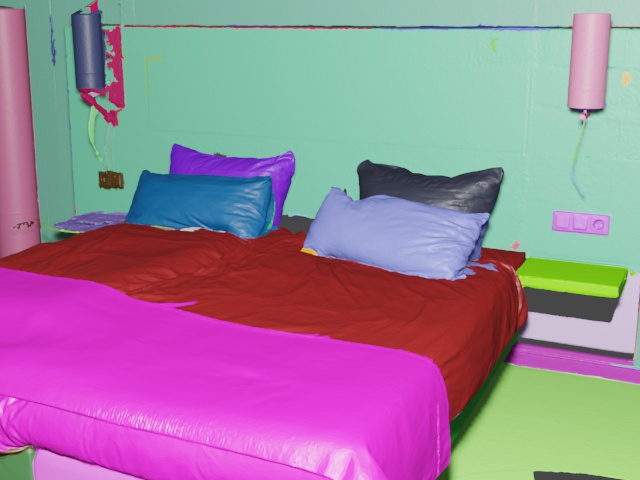}
    \hspace{-4mm}
    &\includegraphics[width=0.19\linewidth]{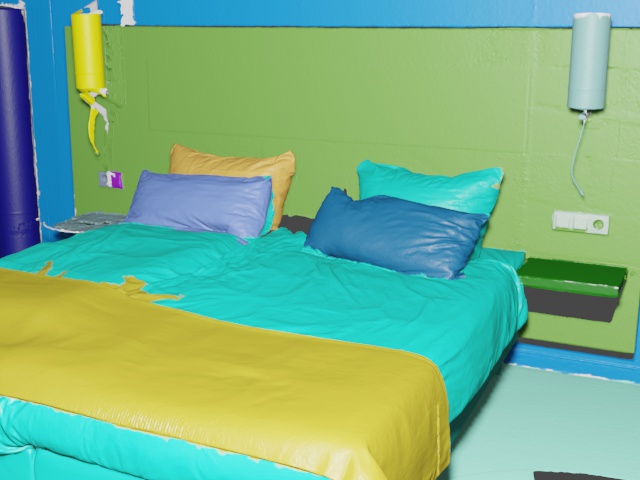} \\

    \small{Input} \hspace{-4mm}
    &\small{SAM3D} \hspace{-4mm} &\small{Mask3D} \hspace{-4mm} &\small{Ours} \hspace{-4mm} & \small{GT} 
    \end{tabular}
    \caption{\small {\textbf{Additional visual results of 3D instance segmentation on ScanNet++ dataset.} }
    }
    
	\label{fig:supp_visual_scannetpp}
\end{figure*}

We show more visual comparisons on ScanNet++ dataset in Fig. \ref{fig:supp_visual_scannetpp}.

\begin{figure*}[t]
    \centering
    \begin{tabular}{cccccc}

    \includegraphics[width=0.19\linewidth]{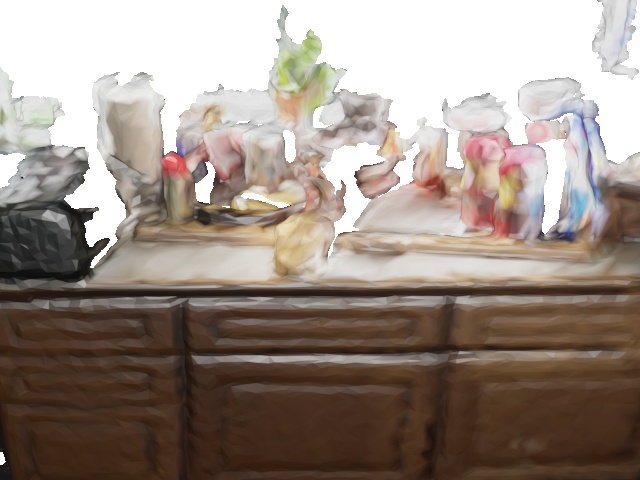}
    \hspace{-4mm}
    &\includegraphics[width=0.19\linewidth]{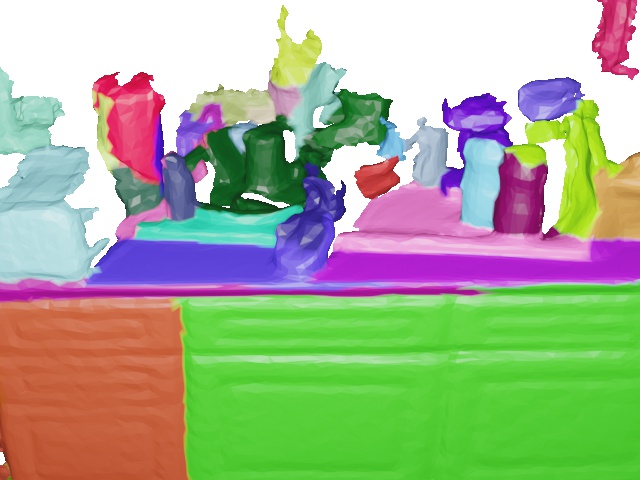}
    \hspace{-4mm}
    &\includegraphics[width=0.19\linewidth]{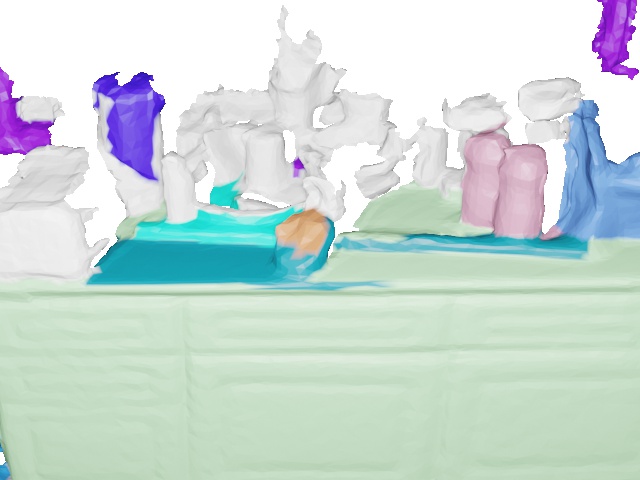}
    \hspace{-4mm}
    &\includegraphics[width=0.19\linewidth]{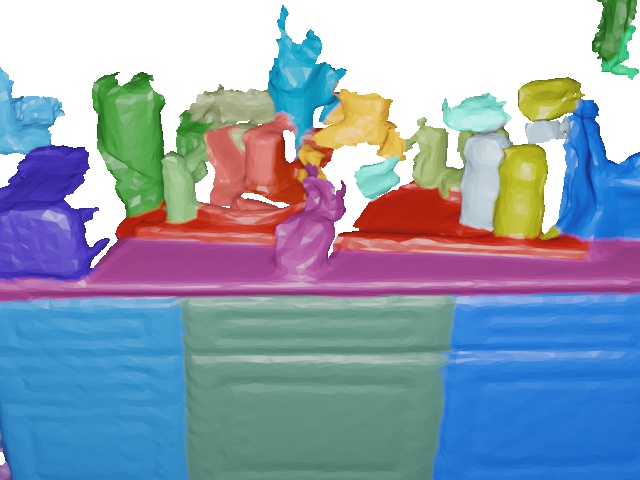}
    \hspace{-4mm}
    &\includegraphics[width=0.19\linewidth]{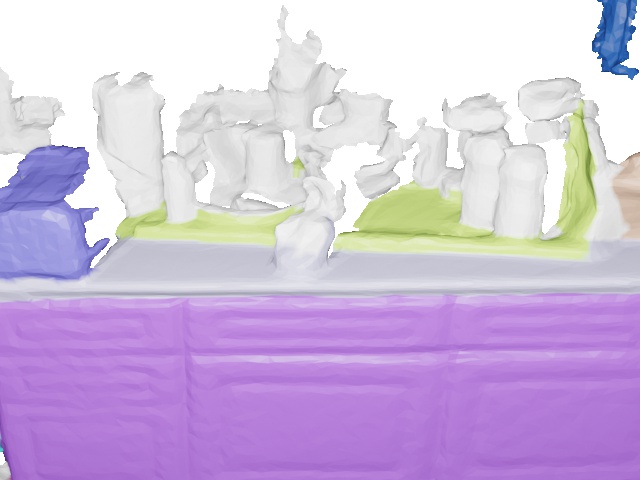} \\

    \includegraphics[width=0.19\linewidth]{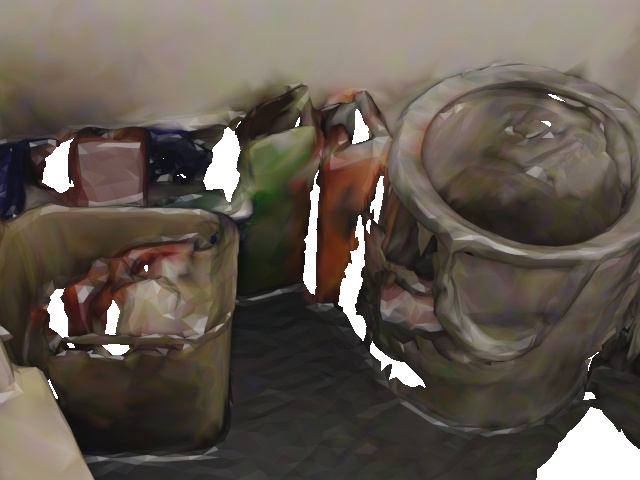}
    \hspace{-4mm}
    &\includegraphics[width=0.19\linewidth]{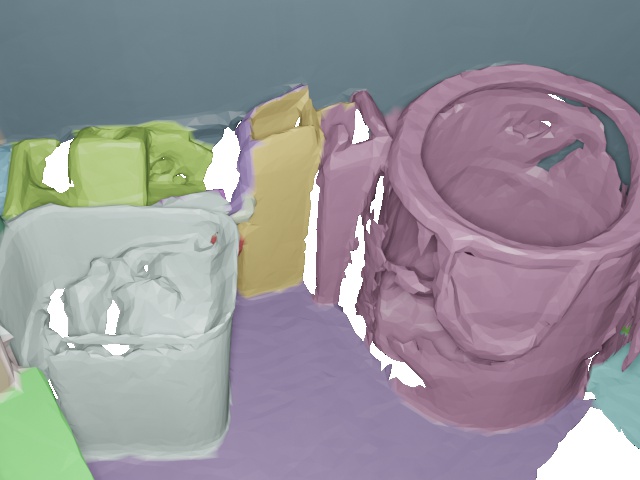}
    \hspace{-4mm}
    &\includegraphics[width=0.19\linewidth]{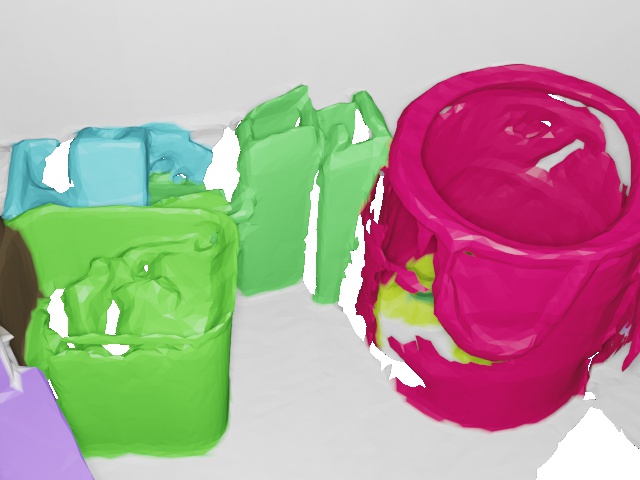}
    \hspace{-4mm}
    &\includegraphics[width=0.19\linewidth]{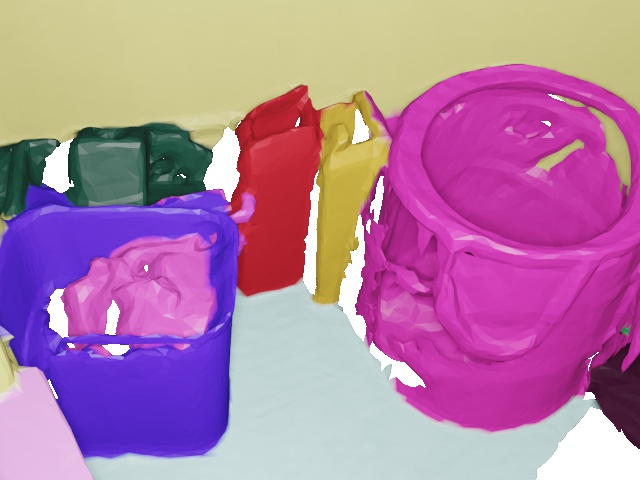}
    \hspace{-4mm}
    &\includegraphics[width=0.19\linewidth]{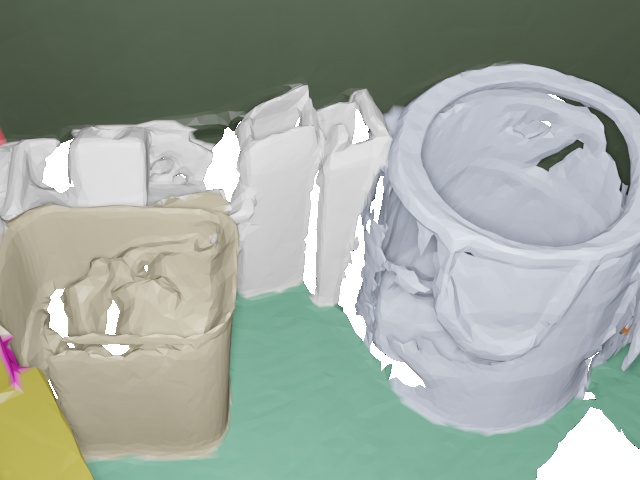} \\

    \includegraphics[width=0.19\linewidth]{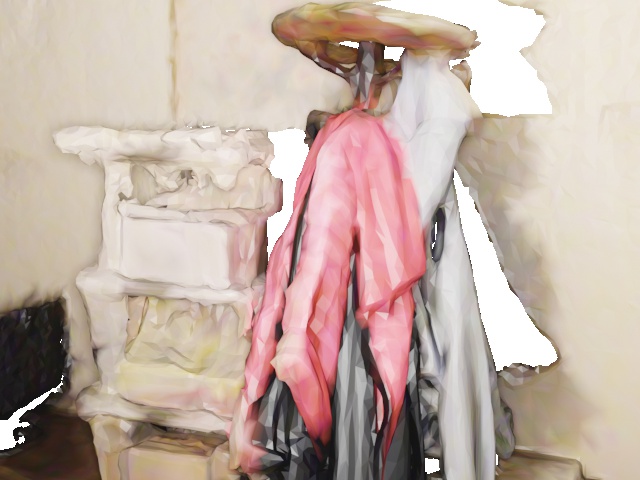}
    \hspace{-4mm}
    &\includegraphics[width=0.19\linewidth]{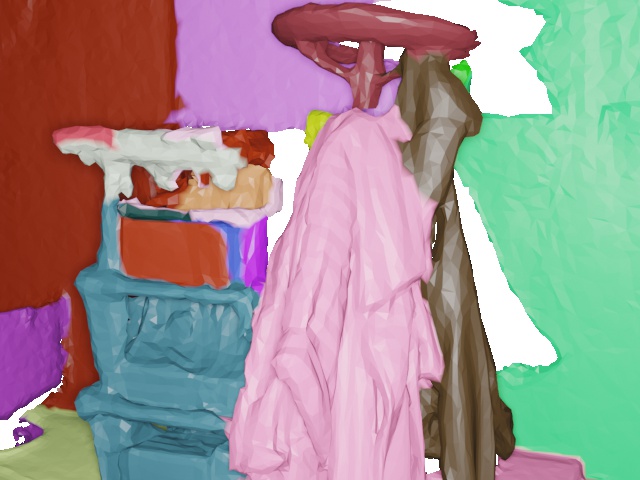}
    \hspace{-4mm}
    &\includegraphics[width=0.19\linewidth]{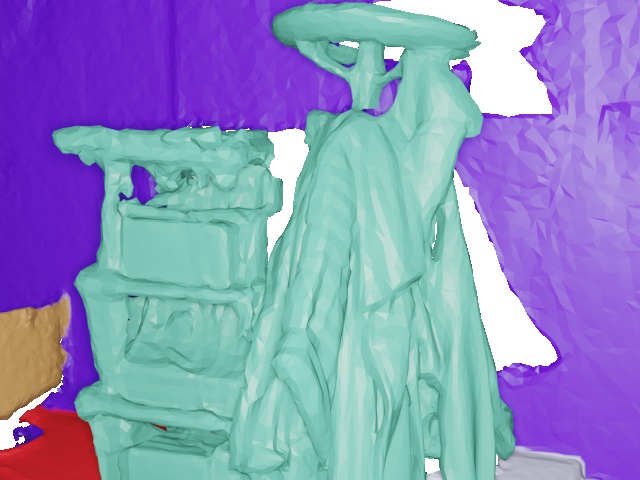}
    \hspace{-4mm}
    &\includegraphics[width=0.19\linewidth]{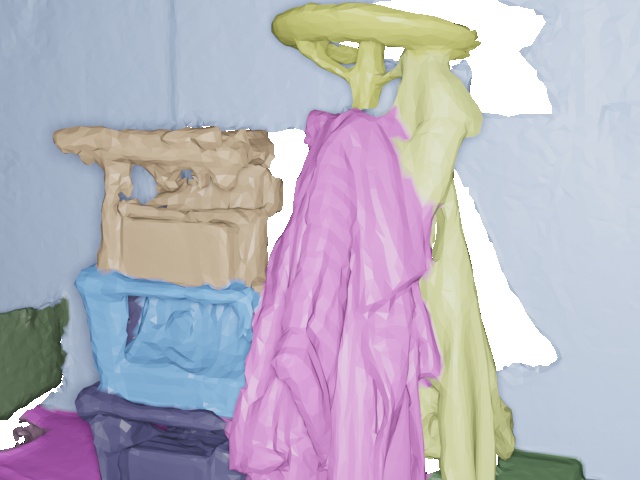}
    \hspace{-4mm}
    &\includegraphics[width=0.19\linewidth]{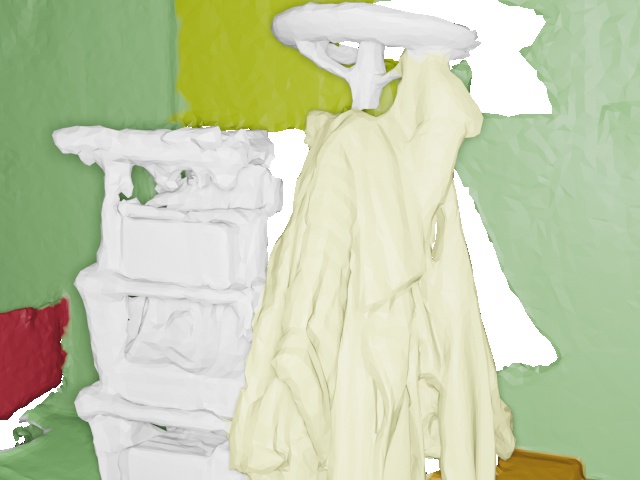} \\

    \includegraphics[width=0.19\linewidth]{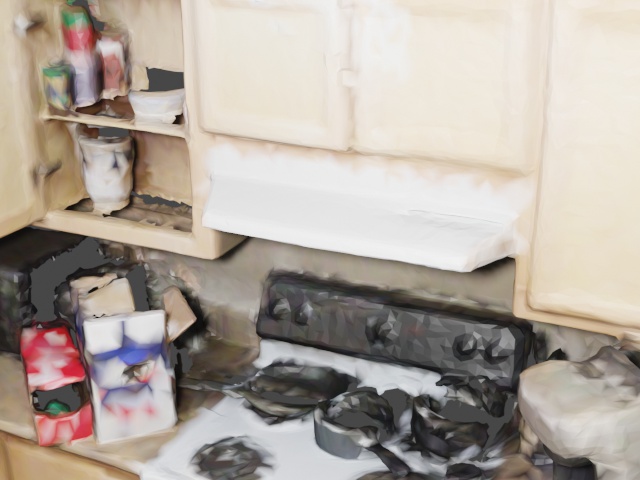}
    \hspace{-4mm}
    &\includegraphics[width=0.19\linewidth]{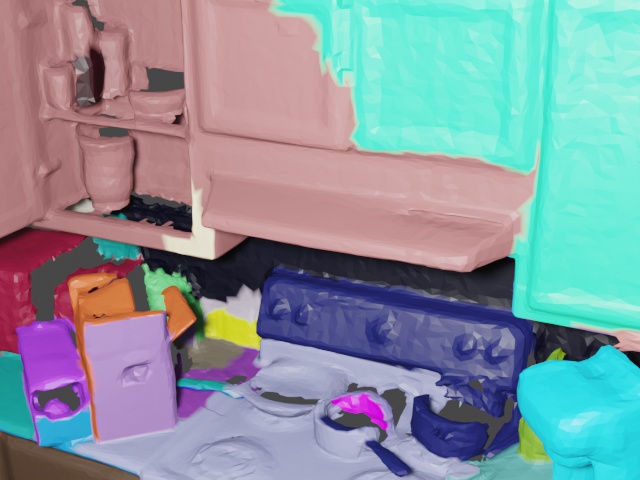}
    \hspace{-4mm}
    &\includegraphics[width=0.19\linewidth]{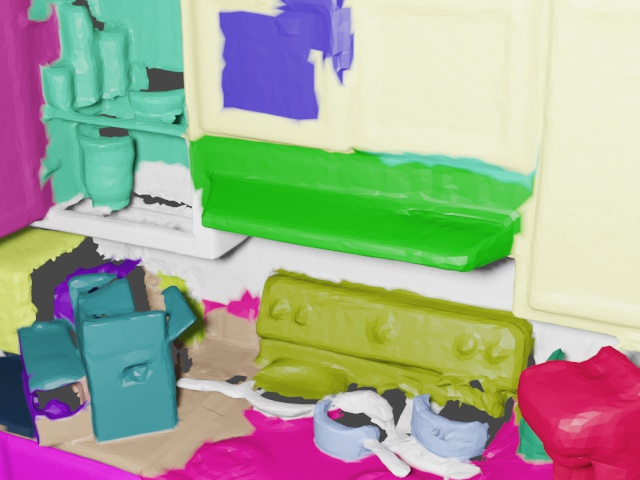}
    \hspace{-4mm}
    &\includegraphics[width=0.19\linewidth]{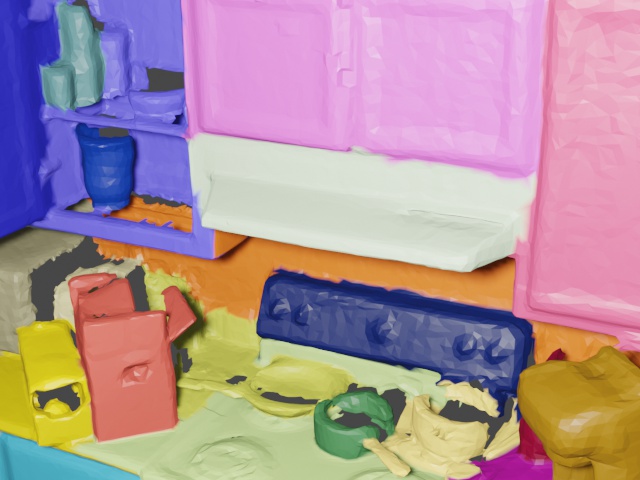}
    \hspace{-4mm}
    &\includegraphics[width=0.19\linewidth]{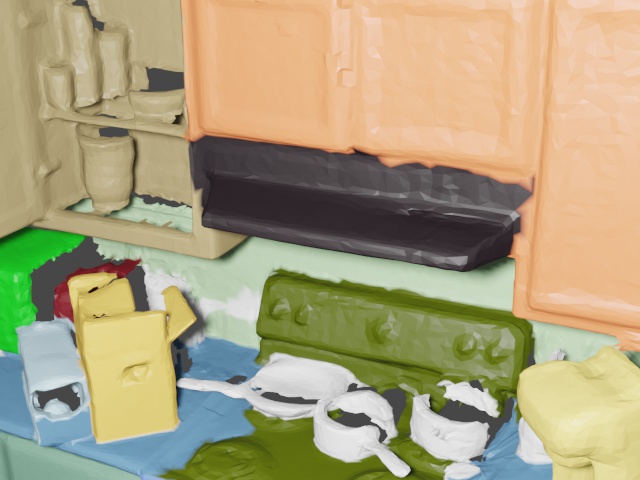} \\

    \includegraphics[width=0.19\linewidth]{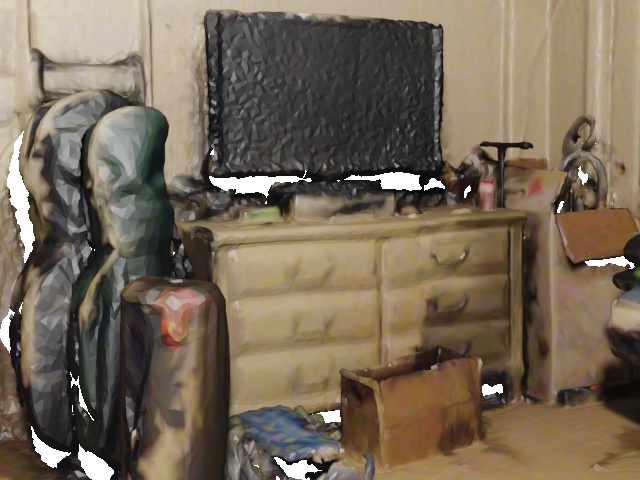}
    \hspace{-4mm}
    &\includegraphics[width=0.19\linewidth]{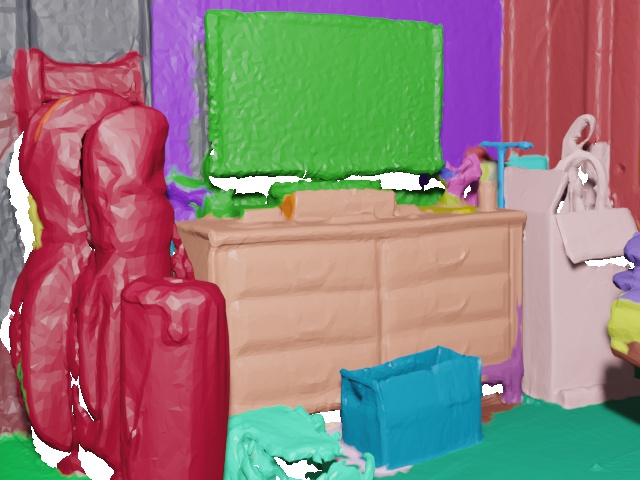}
    \hspace{-4mm}
    &\includegraphics[width=0.19\linewidth]{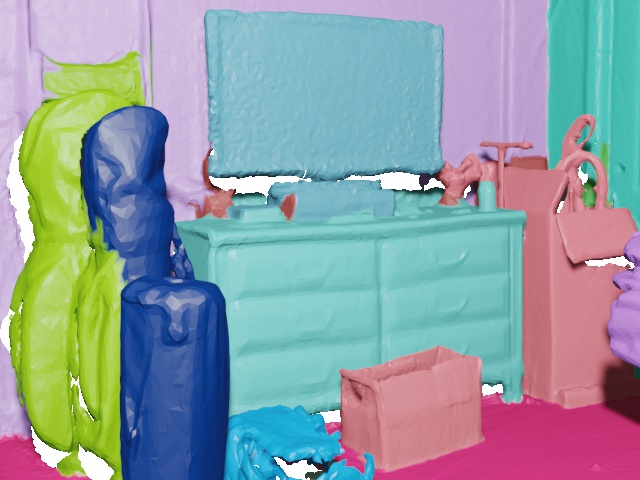}
    \hspace{-4mm}
    &\includegraphics[width=0.19\linewidth]{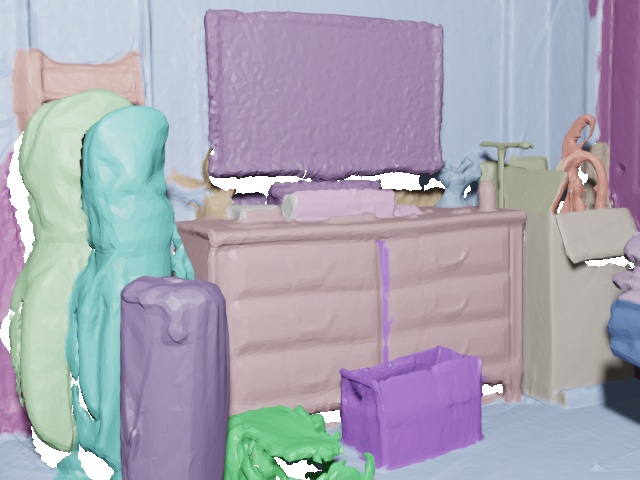}
    \hspace{-4mm}
    &\includegraphics[width=0.19\linewidth]{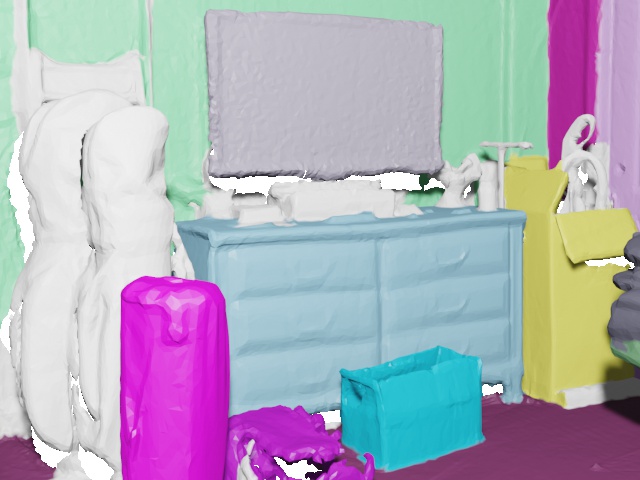} \\

    \includegraphics[width=0.19\linewidth]{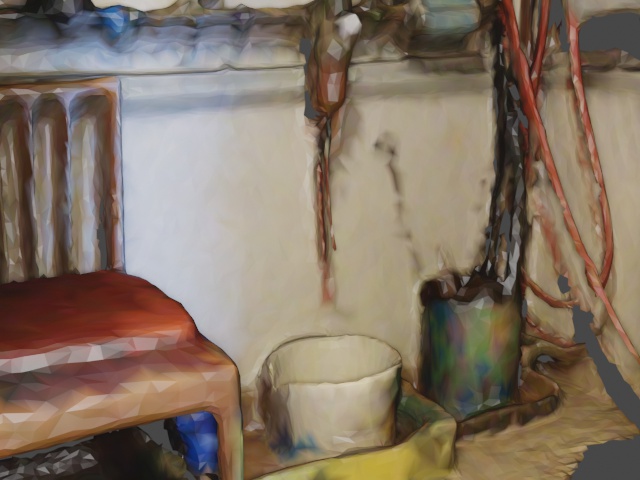}
    \hspace{-4mm}
    &\includegraphics[width=0.19\linewidth]{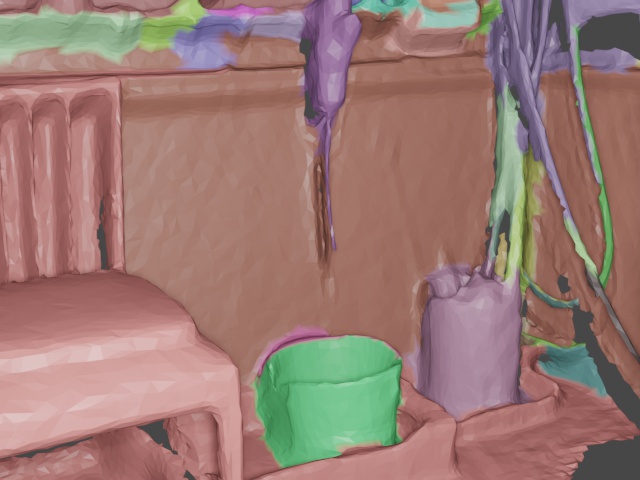}
    \hspace{-4mm}
    &\includegraphics[width=0.19\linewidth]{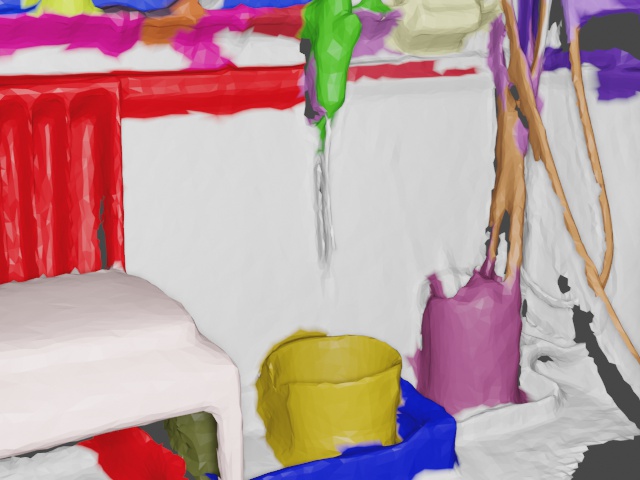}
    \hspace{-4mm}
    &\includegraphics[width=0.19\linewidth]{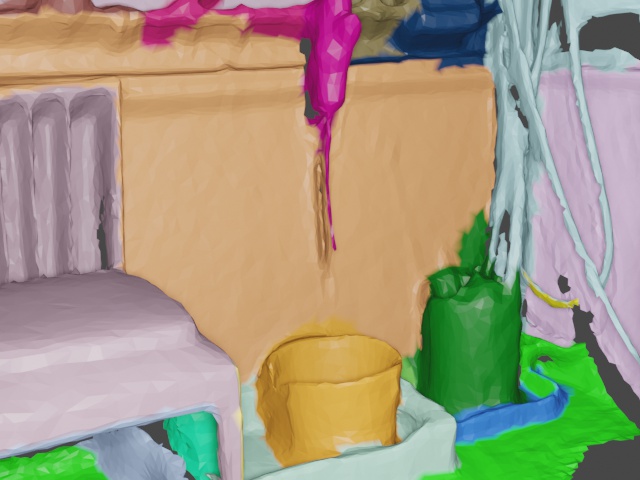}
    \hspace{-4mm}
    &\includegraphics[width=0.19\linewidth]{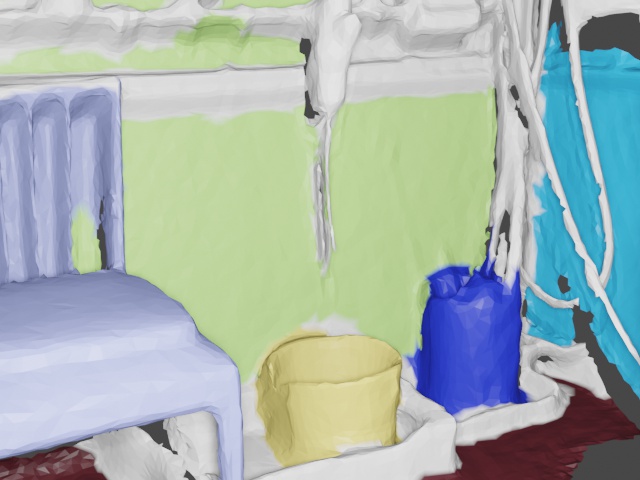} \\

    \includegraphics[width=0.19\linewidth]{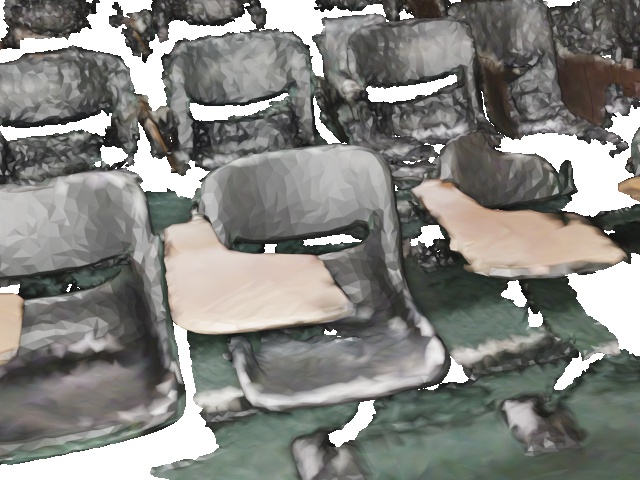}
    \hspace{-4mm}
    &\includegraphics[width=0.19\linewidth]{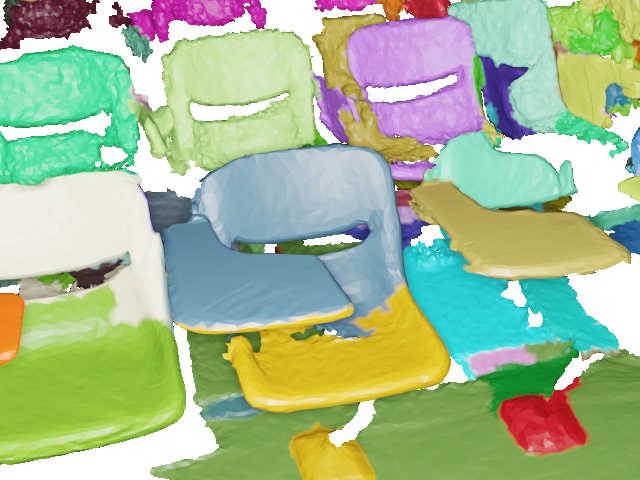}
    \hspace{-4mm}
    &\includegraphics[width=0.19\linewidth]{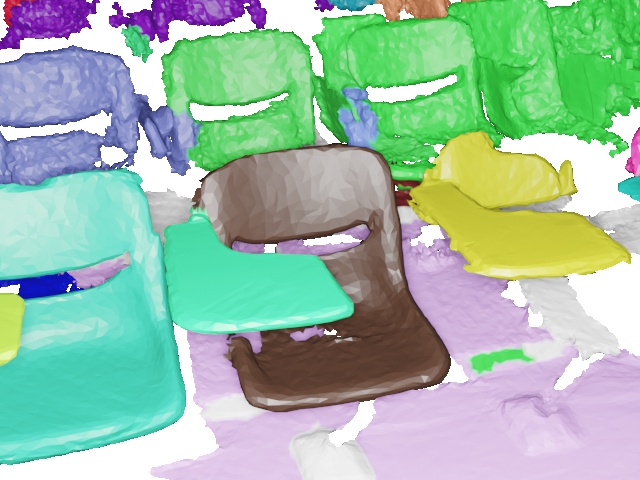}
    \hspace{-4mm}
    &\includegraphics[width=0.19\linewidth]{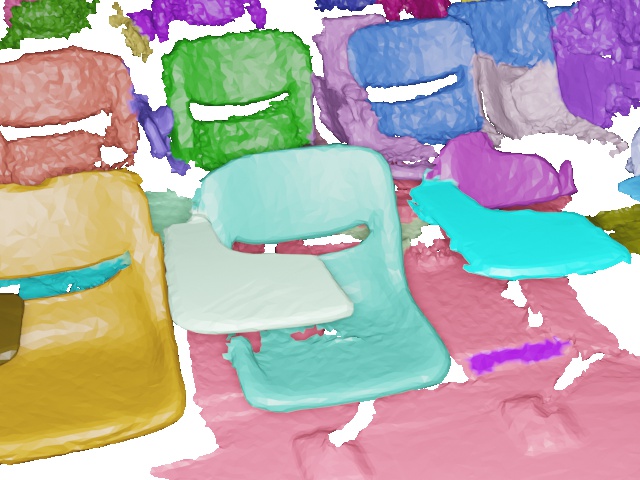}
    \hspace{-4mm}
    &\includegraphics[width=0.19\linewidth]{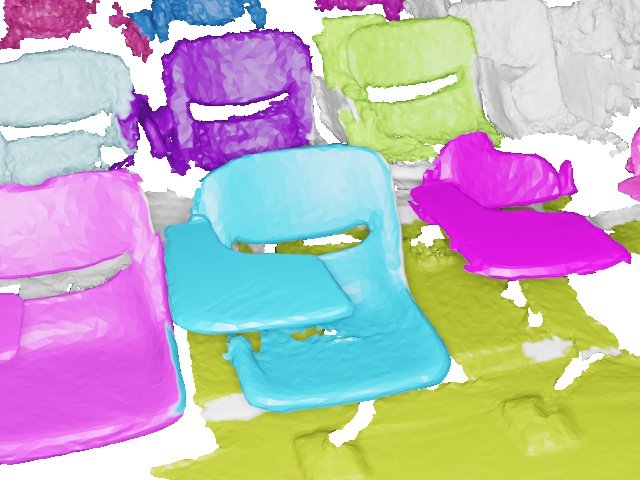} \\

    \includegraphics[width=0.19\linewidth]{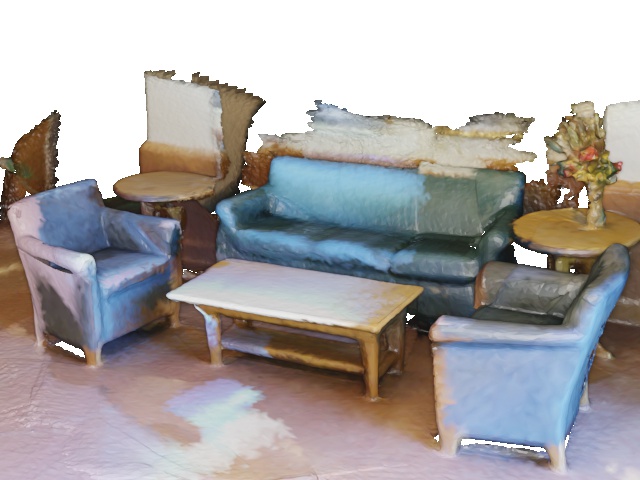}
    \hspace{-4mm}
    &\includegraphics[width=0.19\linewidth]{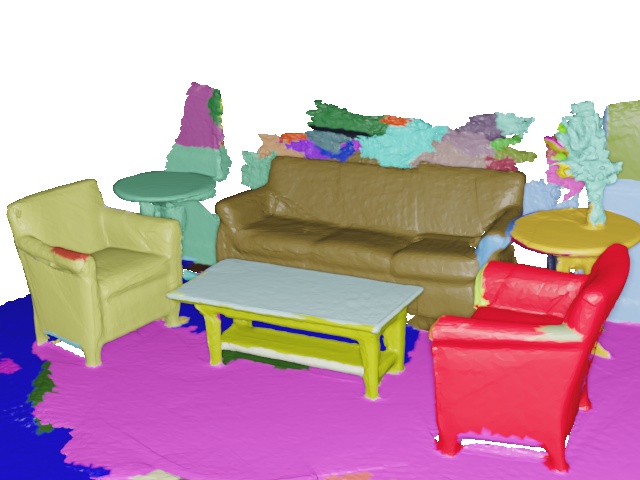}
    \hspace{-4mm}
    &\includegraphics[width=0.19\linewidth]{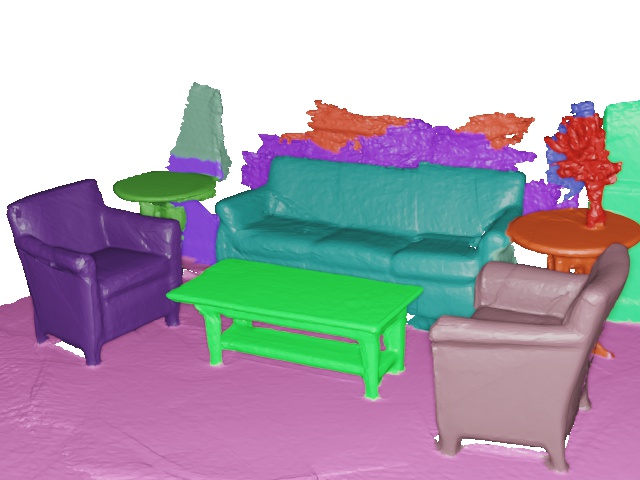}
    \hspace{-4mm}
    &\includegraphics[width=0.19\linewidth]{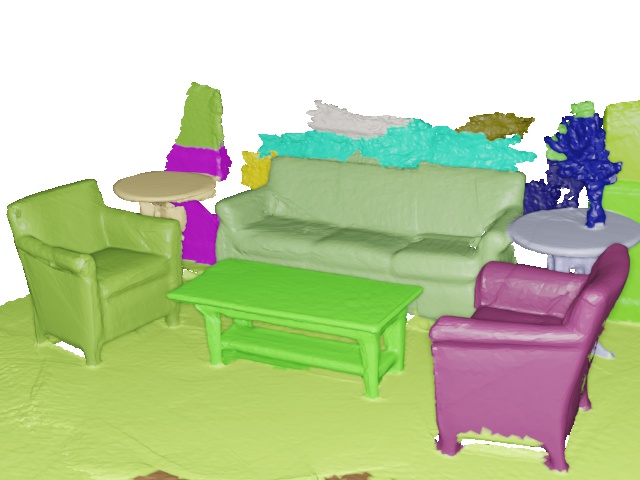}
    \hspace{-4mm}
    &\includegraphics[width=0.19\linewidth]{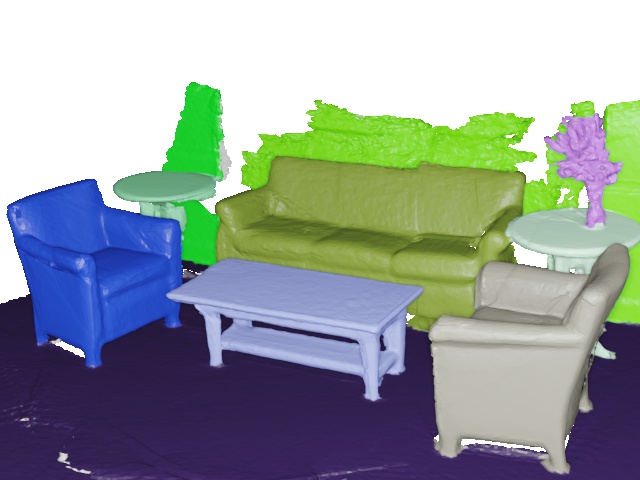} \\

    \small{Input} \hspace{-4mm}
    &\small{SAM3D} \hspace{-4mm} &\small{Mask3D} \hspace{-4mm} &\small{Ours} \hspace{-4mm} & \small{GT} 
    \end{tabular}
    
    \caption{\small {\textbf{Visual results of 3D instance segmentation on ScanNetV2 dataset.}
    }}
    \vspace{-2mm}
	\label{fig:supp_visual_scannet}
\end{figure*}

Visual results on ScanNet dataset are illustrated in Fig. \ref{fig:supp_visual_scannet}. We find that sometimes our method even results in finer and more accurate segmentation masks than the ground truth annotations. See the clutter in the first four rows.

\end{document}